%% file: main.tex
\documentclass{article}

\usepackage[preprint, nonatbib]{neurips_2021}

\usepackage[utf8]{inputenc} 
\usepackage[T1]{fontenc}    
\usepackage{hyperref}       
\usepackage{url}            
\usepackage{booktabs}       
\usepackage{amsfonts}       
\usepackage{nicefrac}       
\usepackage{microtype}      
\usepackage[table]{xcolor}         
\usepackage{subfig}
\usepackage[frozencache,cachedir=.]{minted}

\hypersetup{colorlinks = true, linkcolor = NavyBlue,
            urlcolor  = gray,
            citecolor = NavyBlue,
            anchorcolor = NavyBlue}

\usepackage{rotating}

\usepackage{tabularx}
\usepackage{multirow}
\usepackage{siunitx}
\usepackage{makecell}
\usepackage{amsmath}
\usepackage{cleveref}

\definecolor{Orange}{rgb}{0.9,0.5,0}
\definecolor{NavyBlue}{rgb}{0.1, 0.4, 0.8}
\definecolor{Magenta}{rgb}{0.8, 0.1, 0.6}
\newcommand{\comment}[1]{}

\title{Efficient Training of  Visual Transformers with Small Datasets}

\author{%
  Yahui Liu \\
  University of Trento\\
  Fondazione Bruno Kessler\\
  \texttt{yahui.liu@unitn.it} \\
  \And
  Enver Sangineto \\
  University of Trento\\
  \texttt{enver.sangineto@unitn.it} \\
  \And
  Wei Bi \\
  Tencent AI Lab\\
  \texttt{victoriabi@tencent.com} \\
  \And
  Nicu Sebe\\
  University of Trento\\
  \texttt{niculae.sebe@unitn.it}\\
  \And 
  Bruno Lepri \\
  Fondazione Bruno Kessler\\
  \texttt{lepri@fbk.eu}
  \And
  Marco De Nadai\\
  Fondazione Bruno Kessler\\
  \texttt{work@marcodena.it}
}

\begin{document}

\maketitle

\newcommand{\posimprov}[1]{\scriptsize{(\textcolor{teal}{+#1})}}
\newcommand{\negimprov}[1]{\scriptsize{(\textcolor{brown}{-#1})}}
\newcommand{\negimprovv}[1]{\scriptsize{(\textcolor{brown}{+#1})}}
\newcommand{\arrup}{\textcolor{teal}{$\uparrow$}}
\newcommand{\CC}[1]{\cellcolor{lightgray!#1}}

\input{sections/1-abstract-enver}
\input{sections/2-introduction}

\input{sections/3-related}

\input{sections/3.5-preliminaries}
\input{sections/4-method}

\input{sections/5-experiments-enver}

\input{sections/6-conclusion}

\bibliography{biblio}
\bibliographystyle{plain}

\clearpage

\appendix

\input{supplementary/content}

\end{document}

%% file: sections/1-abstract-enver.tex
\begin{abstract}
Visual Transformers (VTs) 
are emerging as an architectural paradigm  alternative to Convolutional networks (CNNs). Differently from CNNs, VTs can capture global relations between image elements and  they potentially have a larger representation capacity. However, the lack of the typical convolutional  inductive bias 
makes these models
more data hungry than common CNNs. In fact, some local properties of the visual domain which are embedded in the CNN architectural design, in VTs 
should be learned from samples. In this paper, we empirically analyse different VTs, comparing their robustness in a small training set regime, and we show that, despite having a comparable accuracy when trained on ImageNet, their performance on smaller datasets can be largely different. Moreover, we propose an auxiliary  self-supervised task which
can extract additional information from images with only a negligible computational overhead. 
This task  
encourages the VTs to learn  spatial relations within an image and makes the VT training much more robust when training data is scarce.
Our task  is used jointly with the standard (supervised) training and it does not depend on specific architectural choices, thus it can be easily plugged in the existing VTs. Using an extensive evaluation with different VTs and datasets, we show that our method can improve (sometimes dramatically) the final accuracy of  the VTs.
Our code is available at:~\url{https://github.com/yhlleo/VTs-Drloc}.
\end{abstract}

%% file: sections/2-introduction.tex
\section{Introduction}
\label{Introduction}

Visual Transformers (VTs) are progressively emerging architectures in computer vision as an alternative to standard Convolutional Neural Networks (CNNs), and they have already been applied to many tasks, such as image classification \cite{ViT,DeiT,T2T,Swin,CvT,CeiT,LocalViT,CoaT}, object detection \cite{DETR,Deformable-DETR,UP-DETR}, segmentation \cite{strudel2021segmenter}, tracking \cite{TrackFormer},
image generation \cite{TransGAN,GAT} and 3D data processing \cite{PointTransformer}, to mention a few.
These architectures are inspired by the well known Transformer \cite{attention-is-all-you-need}, which is the de facto standard in Natural Language Processing (NLP) 
\cite{devlin-etal-2019-bert,Radford2018ImprovingLU}, 
and one of their appealing properties is the possibility to develop a unified information-processing paradigm for both visual and textual domains.
A pioneering work in this direction is ViT~\cite{ViT}, in which an image is split using a grid of non-overlapping patches, and each patch is linearly projected in the input embedding space, so obtaining a "token". After that, all  the tokens are processed by a series of multi-head attention and feed-forward layers, similarly to how (word) tokens are processed in NLP Transformers.

A clear advantage of VTs is the possibility for the network to use the attention layers to model global relations between tokens, and this is the main difference with respect to CNNs, where the receptive field of the convolutional kernels locally limits the type of relations which can be learned.
However, this increased representation capacity  comes at a price, which is the lack of the typical CNN inductive biases, based on exploiting the locality, the translation invariance and the hierarchical structure of visual information
\cite{Swin,CvT,CeiT}. 
As a result, VTs need a lot of data for training, usually more than what is necessary to standard CNNs
\cite{ViT}. For instance, ViT is trained with JFT-300M
\cite{ViT}, a (proprietary) huge dataset of 303 million (weakly) labeled high-resolution images, and performs worse than ResNets \cite{he2016deep}
with similar capacity when trained on ImageNet-1K ($\sim$ 1.3 million samples~\cite{imagenet}). 
This is likely due to the fact that  ViT needs to learn some local proprieties of the visual data using  more samples than a CNN, while the latter embeds these properties in its architectural design \cite{DBLP:journals/corr/abs-2108-08810}.

To alleviate this problem,  a second generation of VTs has very recently been independently proposed by different groups \cite{T2T,Swin,CvT,CeiT,CoaT,LocalViT,GAT}. A common idea behind these works is to mix convolutional layers with attention layers, in such a way providing a local inductive bias to the VT. These hybrid architectures enjoy the advantages of both paradigms:  attention layers model long-range dependencies, while convolutional operations can emphasize the local properties  of the image content. The empirical results shown in most of these works demonstrate that these second-generation VTs can be trained on ImageNet  outperforming similar-size ResNets on this dataset 
\cite{T2T,Swin,CvT,CeiT,CoaT,LocalViT}. However, it is still not clear what is the behaviour of these networks when trained on medium-small  datasets. In fact, from an application point of view, most of the computer vision tasks cannot rely on (supervised) datasets whose size is comparable with 
(or larger than) ImageNet. 

In this paper, we  compare to each other  different second-generation  VTs 
by either training them from scratch or fine-tuning them on medium-small  datasets, and we empirically show that, despite their ImageNet 
results are basically on par with each other,
their classification accuracy with  smaller datasets largely varies. 
We also compare VTs with same capacity ResNets, and we show that, in most cases, VTs can match the ResNet accuracy when trained with small  datasets.
Moreover, we propose to use an auxiliary self-supervised 
{\em pretext} task and a corresponding
loss function to regularize training in a small training set or few epochs regime.
Specifically, the proposed task is based on (unsupervised) learning the spatial
relations  between the output token embeddings.  
Given an image, we {\em densely} sample random pairs  from the final embedding grid, and, for each pair, we ask the network to guess the corresponding geometric  distance.
To solve this task, the network needs to encode both local and contextual information in each embedding. In fact, without local information, embeddings representing different input image patches cannot be distinguished the one from the others, while, without contextual information (aggregated using the attention layers), the task may be ambiguous.

Our task is  inspired by ELECTRA \cite{ELECTRA},
   in which the (NLP) pretext task 
 is densely defined for each output embedding (Section~\ref{Related}). 
Clark et al.  \cite{ELECTRA}  show that their 
  task is more {\em sample-efficient} than 
 commonly used NLP pretext tasks, and this gain is particularly strong with small-capacity models or relatively smaller training sets. 
Similarly, we exploit the fact that an image is represented by a VT using multiple token embeddings, and we use their relative distances
to define a localization task over a subset of all the possible embedding pairs.
This way, {\em for a single image forward pass}, we can compare many embedding pairs  with each other,
and average our localization loss over all of them.
 Thus, our task is drastically different from those multi-crop strategies proposed, for instance, in SwAV~\cite{SwAV}, which need to independently forward each input patch through the network. 
 Moreover, differently from "ordering" based tasks \cite{noroozi2016unsupervised}, 
 we can define  pairwise distances  on a large grid without 
 modeling all the possible permutations (more details in Section~\ref{Related}).

Since our auxiliary task is self-supervised,
our {\em dense relative localization} loss ($\mathcal{L}_{drloc}$) does not require additional annotation, and we use it jointly with the standard (supervised) cross-entropy as a regularization of the VT training.
$\mathcal{L}_{drloc}$ is very  easy-to-be-reproduced and, despite this simplicity, it can largely boost the accuracy  of the VTs,
especially when the VT is either trained from scratch on a small dataset, or fine-tuned on a dataset with a large domain-shift with respect to the pretraining ImageNet dataset.
In our empirical analysis, based on
 different training scenarios, a variable amount of training  data and  different  VT architectures,
$\mathcal{L}_{drloc}$ has {\em always} improved the results  of the tested baselines, 
sometimes boosting the final accuracy of tens of points (and up to 45 points).

In summary, 
our main contributions are:

\begin{enumerate}
\item
 We empirically compare to each other different VTs, showing that their behaviour  largely differs when trained with small datasets or few training epochs.
\item
We propose a  relative localization  auxiliary  task for VT training regularization.
\item
 Using an extensive empirical 
analysis,  we show that this task is  beneficial to speed-up training and improve the generalization 
ability of different VTs, independently of their specific architectural design or application task.
\end{enumerate}

%% file: sections/3-related.tex
\section{Related work}
\label{Related}

In this section, we briefly review previous work related to both VTs and self-supervised learning.

{\bf Visual Transformers.} Despite some previous work in which attention is used inside the convolutional layers of a CNN \cite{non-local-net,DBLP:conf/iccv/HuZXL19}, 
the first fully-transformer architectures for vision are  iGPT \cite{iGPT} and ViT \cite{ViT}. The former is trained using a "masked-pixel" self-supervised approach, similar in spirit to the common masked-word task used, for instance, in BERT \cite{devlin-etal-2019-bert} and in GPT \cite{Radford2018ImprovingLU} (see below). On the other hand, ViT is  trained in a supervised way, using a special "class token" and a classification head attached to the final embedding of this token. Both methods are computationally expensive and, despite their very good results when trained on huge datasets, they underperform ResNet architectures when trained from scratch using only ImageNet-1K \cite{ViT,iGPT}. VideoBERT \cite{VideoBERT} is conceptually similar to iGPT, but, rather than using pixels as tokens,  each frame of a video is holistically represented by a feature vector, which is quantized using an off-the-shelf pretrained video classification model.
DeiT \cite{DeiT} trains ViT using distillation information provided by a pretrained CNN.

The success of ViT has attracted a lot of interest  in the computer vision community, and different variants of this architecture have been recently used in many tasks \cite{DeiT,strudel2021segmenter,TransGAN,DBLP:journals/corr/abs-2104-02057}. However, as  mentioned in Section~\ref{Introduction}, 
the lack of the typical CNN inductive biases in ViT, makes this model difficult to train without using (very) large datasets. For this reason, very recently, a second-generation of VTs has focused on hybrid architectures, in which convolutions are used jointly with long-range attention layers \cite{T2T,Swin,CvT,CeiT,CoaT,LocalViT,GAT}. The common idea behind all these works is that the sequence of the individual token embeddings can be shaped/reshaped in a geometric grid, in which the position of each embedding vector corresponds to a fixed location in the input image. Given this geometric layout of the embeddings, 
convolutional layers can be applied to neighboring embeddings, so encouraging the network to focus on local properties of the image. The main difference among these works concerns {\em where} the convolutional operation is applied (e.g., only in the initial representations \cite{T2T} or in all the layers \cite{Swin,CvT,CeiT,CoaT,LocalViT}, in the token to query/key/value projections \cite{CvT} or in the forward-layers \cite{CeiT,LocalViT,GAT}, etc.).
In most of the experiments of this paper, 
we use three state-of-the-art
second-generation  VTs for which there is a public  implementation: T2T \cite{T2T}, Swin \cite{Swin} and CvT \cite{CvT}). For each of them, we select the model whose number of parameters is comparable with a ResNet-50 \cite{he2016deep} (more details in Section~\ref{Preliminaries} and Section~\ref{sec:experiments}). We do not modify the native architectures because the goal of this work is to propose a pretext task and a loss function which can be easily plugged in existing VTs.

Similarly to the original Transformer \cite{attention-is-all-you-need}, in ViT, an (absolute) {\em positional embedding} is added to the representation of the input tokens. In Transformer networks, positional embedding is used to provide information about the  token order, since both the attention and the (individual token based) feed-forward layers are permutation invariant. In  \cite{Swin,CoaT}, {\em relative} 
positional embedding \cite{DBLP:conf/naacl/ShawUV18} is used, where the  position of each token is represented relatively to the others. 
Generally speaking, positional embedding is a representation of the token position which is {\em provided as input} to the network. Conversely, our relative localization loss exploits the relative positions (of the final VT embeddings) as a {\em pretext task} to  extract additional information  without manual supervision.

{\bf Self-supervised learning.} Reviewing the vast self-supervised learning literature  is out of the scope of this paper. However, we briefly mention that self-supervised learning was first successfully applied in NLP, as a means to get supervision from text by replacing costly manual annotations with {\em pretext} tasks \cite{DBLP:journals/corr/abs-1301-3781,DBLP:conf/nips/MikolovSCCD13}. A typical NLP pretext task consists in masking a word in an input sentence and asking the network to guess which is the masked token
\cite{DBLP:journals/corr/abs-1301-3781,DBLP:conf/nips/MikolovSCCD13,devlin-etal-2019-bert,Radford2018ImprovingLU}. 
ELECTRA \cite{ELECTRA} is a {\em sample-efficient} language model in which the masked-token pretext task is replaced by a discriminative task defined over all the tokens of the input sentence. Our work is inspired by this method, since we propose a pretext task which can be efficiently computed by densely sampling the final VT embeddings.
However, while the densely supervised ELECTRA task is obtained by randomly replacing  (word) tokens and using a pre-trained BERT model to generate plausible replacements, 
we do not need a pre-trained model and we do not replace input tokens, being our task based on predicting the inter-token geometric distances.
In fact,  in NLP tasks, tokens are discrete and limited (e.g., the set of words of a specific-language dictionary), while image patches are “continuous” and highly variable, hence a replacement-based task is hard to use in a vision scenario.

In computer vision, common pretext tasks with still images are based on extracting two different views from the same image (e.g., two different crops) and then considering these as a pair of {\em positive} images, likely sharing the same semantic content
\cite{simclr}. Most current self-supervised computer vision approaches can be  categorised in contrastive learning \cite{CPC,DIM,simclr,MoCo,tian2019contrastive,wang2020understanding,little-friends}, clustering methods 
\cite{NIPS2016_65fc52ed,DBLP:journals/corr/abs-1903-12355,IIC,DeepClustering,DBLP:conf/iclr/AsanoRV20a,DBLP:conf/eccv/GansbekeVGPG20,SwAV,DINO}, 
asymmetric networks   \cite{byol,simsiam} and feature-decorrelation methods \cite{w-mse,barlow,VICReg,hua2021feature}. 
While the aforementioned approaches are all based on ResNets, very recently, both \cite{DBLP:journals/corr/abs-2104-02057} and \cite{DINO} have empirically tested some of these methods with a ViT architecture \cite{ViT}. 

One important difference of our proposal with respect to previous work, is that we do not propose a fully-self-supervised method, but we rather use  self-supervision jointly with standard supervision (i.e., image labels) in order to regularize VT training, hence
 our framework is a {\em multi-task learning} approach \cite{DBLP:journals/corr/abs-2009-09796}. 
Moreover, our dense relative localization loss is not based on positive pairs, and we do {\em not} use multiple views of the same image in the current batch, thus our method can be used with standard (supervised) data-augmentation techniques. Specifically, our pretext task is based on predicting the relative positions of pairs of tokens extracted from the same image. 

Previous work using localization for self-supervision is based on predicting the input image rotation \cite{RotNet} or the relative position of {\em adjacent patches} extracted from the same image
\cite{doersch2015unsupervised,noroozi2016unsupervised,noroozi2017representation,misra2019selfsupervised}. For instance, 
in \cite{noroozi2016unsupervised}, the network should predict the correct permutation of a grid of $3 \times 3$ patches
(in NLP, a similar, permutation based pretext task, is {\em deshuffling} \cite{DBLP:journals/jmlr/RaffelSRLNMZLL20}).
In contrast,  we do not need to extract multiple patches from the same input image, since we can efficiently use the final token embeddings (thus, we need a {\em single} forward and backward pass per image). Moreover, differently from previous work based on localization pretext tasks, our loss is {\em densely} computed between many random pairs of (non necessarily adjacent) token embeddings.
For instance, a trivial extension of the ordering task proposed in \cite{noroozi2016unsupervised} using a grid of $7 \times 7$ patches would lead to 
49! possible  permutations, which becomes intractable if  modeled  as a classification task.
Finally, in \cite{UP-DETR}, the position of a random query patch is used for the  self-supervised training  of a  transformer-based object detector
\cite{DETR}. However, the localization loss used in \cite{UP-DETR} is specific for the final task (object localization) and the specific DETR architecture \cite{DETR}, while our loss is generic and can be plugged in any VT.

%% file: sections/3.5-preliminaries.tex
\begin{figure}[ht]
    \centering
    \includegraphics[width=\linewidth]{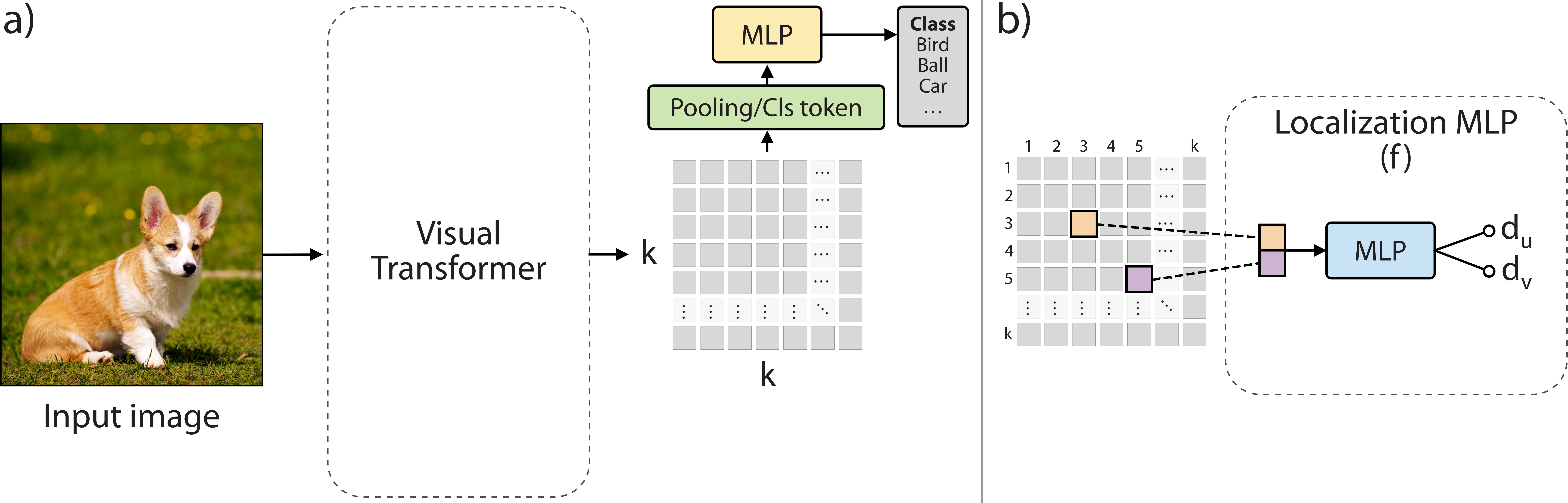}
    \caption{A schematic representation of the VT architecture. (a) A typical second-generation VT. (b) Our localization MLP which takes as input (concatenated) pairs of final token embeddings.}
    \label{fig:my_label}
    \vspace{-1em}
\end{figure}

\section{Preliminaries}
\label{Preliminaries}

A typical VT network takes as input an image  split in a grid of (possibly overlapping) $K \times K$ patches. Each patch is  projected in the input embedding space, obtaining a set of  $K \times K$ input {\em tokens}. A VT is based on the  typical Transformer
multi-attention layers
\cite{attention-is-all-you-need}, which model pairwise relations over the token intermediate representations. 
Differently from a pure Transformer 
\cite{attention-is-all-you-need}, the hybrid architectures mentioned in Section~\ref{Introduction}-\ref{Related}
usually shape or reshape the sequence of these token embeddings in a spatial grid, which makes it possible to apply convolutional operations over a small set of neighboring  token embeddings. Using convolutions with a stride greater than 1 and/or pooling operations, the resolution of the initial $K \times K$ token grid can possibly be reduced, thus simulating the hierarchical structure of a CNN. We assume that the final embedding  grid has a resolution of $k \times k$ (where, usually, $k \leq K$), see Fig.~\ref{fig:my_label} (a).

The final $k \times k$ grid of embeddings represents the input image and it is used for the discriminative task.
For instance, some methods include an additional "class token" which collects contextual information over the whole grid \cite{ViT,T2T,CvT,CeiT,CoaT,LocalViT}, while others \cite{Swin} apply an average global pooling  over the final grid to get a compact  representation of the whole image. Finally, a standard, small MLP  head takes as input the whole image representation and it outputs a posterior distribution over the set of the target classes (Fig.~\ref{fig:my_label} (a)). The  VT is trained using a standard cross-entropy loss
($\mathcal{L}_{ce}$), computed using these posteriors and the image ground-truth labels.

When we plug our relative localization loss (Section~\ref{Method}) in an existing VT, we 
 always use the native VT architecture of each tested method, without any change apart from the dedicated localization MLP (see Section~\ref{Method}). For instance, we use the class token when available, or the average pooling layer when it is not, and on top of these we use the cross-entropy loss.
We also keep the positional embedding (Section~\ref{Related}) for those VTs which use it (see Section~\ref{Discussion} for a discussion about this choice).
 The only architectural change we do is to downsample the final embedding grid of T2T \cite{T2T} and CvT \cite{CvT} to make them of the same size as that used in Swin \cite{SwAV}. Specifically, in Swin, the final grid has a resolution of $7 \times 7$ ($k = 7$), while, in T2T and in CvT, it is $14 \times 14$. Thus, in T2T and in CvT, we use a $2 \times 2$ average pooling ({\em without} learnable parameters) and we get a final $7 \times 7$ grid for all the three tested architectures. This pooling operation is motivated in Section~\ref{Discussion}, and
 it is used only together with our localization task (it does not affect the posterior computed by the classification MLP).
 Finally, note that T2T uses convolutional operations only in the input stage, and it outputs a sequence of $14 \times 14 = 196$ embeddings,
 corresponding to its $14 \times 14$ input grid. In this case, we first reshape the sequence and then we use pooling.
 In the Supplementary Material, we show additional experiments with a ViT architecture \cite{ViT}, in which we adopt
 the same reshaping and pooling strategy.

%% file: sections/4-method.tex
\section{Dense relative localization task}
\label{Method}

The goal of our regularization task is to encourage the VT to learn spatial information without using additional manual annotations.
We achieve this by {\em densely} sampling  {\em multiple} embedding pairs {\em for each image} and asking the network to guess their relative distances.
In more detail, given 
an image $x$, we denote  its corresponding $k \times k$ grid of final embeddings (Section~\ref{Preliminaries}),  as 
$G_x = \{ \mathbf{e}_{i,j} \}_{1 \leq i,j \leq k}$, where $\mathbf{e}_{i,j} \in \mathbb{R}^d$, and $d$ is the dimension of the embedding space.
For each $G_x$, we randomly sample multiple pairs of embeddings and, for each pair $(\mathbf{e}_{i,j}, \mathbf{e}_{p,h})$, we compute the 2D 
normalized target translation offset  $(t_u,t_v)^T$, 
where:
\begin{equation}
\label{eq.trans-offset}
    t_u = \frac{|i-p|}{k}, \quad t_v = \frac{|j-h|}{k}, \quad (t_u,t_v)^T \in [ 0,  1 ]^2.
\end{equation}
The selected embedding vectors $\mathbf{e}_{i,j}$ and $\mathbf{e}_{p,h}$ are concatenated and input to a small MLP ($f$), 
with two hidden layers 
and two output neurons, one per  spatial dimension (Fig.~\ref{fig:my_label} (b)), which predicts the relative distance between  position $(i,j)$ and position $(p,h)$ on the grid. Let $(d_u,d_v)^T = f(\mathbf{e}_{i,j}, \mathbf{e}_{p,h})^T$. 
Given a mini-batch $B$ of $n$ images, our {\em dense relative localization loss} is:
\begin{equation}
\label{eq.drloc-loss}
\mathcal{L}_{drloc} = 
\sum_{x \in B} 
\mathbb{E}_{(\mathbf{e}_{i,j}, \mathbf{e}_{p,h}) \sim G_x}
    [ | (t_u,t_v)^T - (d_u,d_v)^T |_1 ].
\end{equation}
\noindent
In Eq.~\ref{eq.drloc-loss}, for each image $x$, the expectation is computed by sampling uniformly at random 
$m$  pairs $(\mathbf{e}_{i,j}, \mathbf{e}_{p,h})$ in $G_x$, and
 averaging the $L_1$ loss between the corresponding $(t_u,t_v)^T$ and $(d_u,d_v)^T$.

$\mathcal{L}_{drloc}$ is added to the standard cross-entropy loss ($\mathcal{L}_{ce}$) of each native VT (Section~\ref{Preliminaries}). The final loss is: $\mathcal{L}_{tot} = \mathcal{L}_{ce} + \lambda \mathcal{L}_{drloc}$. We use $\lambda = 0.1$ in  all the experiments with both T2T and CvT, and 
$\lambda = 0.5$ in case of Swin. 
Note that the same pairwise localization task can be associated with slightly different loss formulations. 
In the Supplementary Material we  present some of these variants and 
we compare them empirically with each other.

\subsection{Discussion}
\label{Discussion}

Intuitively, $\mathcal{L}_{drloc}$ transforms the relative positional embedding 
(Section~\ref{Related}), used, for instance, in Swin \cite{Swin}, in a pretext task, asking the network to guess which is the relative distance of a random subset of all the possible token pairs. 
Thus a question arises: is the relative positional embedding used in some VTs sufficient for the localization MLP ($f$) to solve the localization task?
The experiments presented in Section~\ref{scratch}-\ref{finetuning} show that, when we plug $\mathcal{L}_{drloc}$ on CvT, in which {\em no kind} of positional embedding is used \cite{CvT}, the relative accuracy boost is usually {\em smaller} than in case of Swin, confirming that the relative positional embedding, used in the latter, is not sufficient to make our task trivial. We further analyze this point in the Supplementary Material.

In Section~\ref{Preliminaries}, we mentioned that, in case  
of T2T and CvT,
we average-pool the final grid and we obtain a   $7 \times 7$ grid $G_x$. 
In fact, in preliminary experiments with both T2T and CvT at their original $14 \times 14$ resolution, we observed a very slow convergence of $\mathcal{L}_{drloc}$. We presume this is due to the fact that, with a finer grid, the localization task is harder. This  slows down the convergence of $f$, and it likely generates noisy gradients which are backpropagated through the whole VT (see also the Supplementary Material).
We leave this for future investigation and, in the rest of this article, we always assume that our pretext task is computed with a $7 \times 7$ grid $G_x$.

\begin{table}[!ht]
	\small 
	\centering
	\caption{The size of the datasets  used in our empirical analysis.}
	\renewcommand{\arraystretch}{1.0}
	\setlength{\tabcolsep}{6pt}
	\input{tables/datasets}
	\vspace{-1.5em}
	\label{tab:datasets}
\end{table}

%% file: tables/datasets.tex
	\begin{tabular}{llrrr}
			\toprule
			\multicolumn{2}{l}{\textbf{Dataset}}  & \textbf{Train size} & \textbf{Test size}  & \textbf{Classes}\\
			\midrule
			\multicolumn{2}{l}{ImageNet-1K~\cite{imagenet}} & 1,281,167 & 100,000 & 1000 \\
			\multicolumn{2}{l}{ImageNet-100~\cite{tian2019contrastive}} & 126,689 & 5,000 & 100 \\
			\midrule
			\multicolumn{2}{l}{CIFAR-10~\cite{cifar}} & 50,000 & 10,000 & 10 \\
			\multicolumn{2}{l}{CIFAR-100~\cite{cifar}}  & 50,000 & 10,000 & 100 \\
			\multicolumn{2}{l}{Oxford Flowers102~\cite{flower}} & 2,040 & 6,149 & 102 \\
			\multicolumn{2}{l}{SVHN~\cite{netzer2011reading}} & 73,257 & 26,032 & 10 \\
			\midrule
			\multirow{6}{*}{\rotatebox{90}{\hspace*{-2pt}DomainNet}} & ClipArt & 33,525 & 14,604 & \multirow{6}{*}{345} \\
			& Infograph & 36,023 & 15,582  \\
			& Painting & 50,416 & 21,850 \\
			& Quickdraw~ & 120,750 & 51,750  \\
			& Real & 120,906 & 52,041 \\
			& Sketch & 48,212 & 20,916 \\
			\bottomrule
		\end{tabular}

%% file: sections/5-experiments-enver.tex
\section{Experiments}
\label{sec:experiments}

All the  experiments presented in this section are based on image classification tasks, while in 
the Supplementary Material 
we also show object detection, instance segmentation and semantic segmentation tasks. In this section we use 11 different datasets:  
ImageNet-100 (IN-100)
\cite{tian2019contrastive, wang2020understanding}, which is a subset of 100 classes of ImageNet-1K~\cite{imagenet};
CIFAR-10 and CIFAR-100~\cite{cifar}, Oxford Flowers102~\cite{flower} and SVHN~\cite{netzer2011reading}, which are four widely used computer vision datasets; and the six datasets of  DomainNet~\cite{peng2019moment}, a benchmark commonly used for domain adaptation tasks. We chose the latter 
because of the  large domain-shift between some of its datasets and ImageNet-1K, which makes the  
 fine-tuning experiments non-trivial.
Tab.~\ref{tab:datasets}  shows the size of each dataset.

We used, when available,  the official VT code (for T2T \cite{T2T} 
and Swin \cite{Swin}) 
and a publicly available implementation of
CvT \cite{CvT}\footnote{\url{https://github.com/lucidrains/vit-pytorch}}. 
In the fine-tuning experiments (Section~\ref{finetuning}), we use only T2T and Swin 
because of the lack of publicly available ImageNet pre-trained CvT networks. 
For each of the three baselines, 
we chose a model of comparable size to ResNet-50 (25M parameters): see Tab.~\ref{tab:results-IN-100} for more details.
In the Supplementary Material, we show additional results obtained with larger models (ViT-B \cite{ViT}),
larger datasets (e.g., ImageNet-1K) and more training epochs.
When we plug our loss on one of the adopted baselines, we follow Section~\ref{Method}, {\em keeping unchanged} the VT architecture apart from our localization MLP ($f$). Moreover, in all the experiments, 
we train the baselines, both with and without our localization loss, 
using the same data-augmentation protocol for all the models, 
and we use the VT-specific
hyper-parameter configuration suggested by the authors of each VT. 
We do {\em not} tune the VT-specific hyperparameters when we use our loss and we keep fixed the values of $m$ and $\lambda$ (Section~\ref{Ablation}) in all the experiments. 
We train each model using 8  V100 32GB GPUs.

\subsection{Ablation study}
\label{Ablation}

In Tab.~\ref{tab:ablation} (a)
we analyze the impact on the accuracy of different values of $m$ (the total number of embedding pairs used per image, see Section~\ref{Method}). Since we use the same grid resolution for all the VTs (i.e., $7 \times 7$, Section~\ref{Preliminaries}),
also the maximum number of possible embeddings per image is the same for all the VTs ($k^2 = 49$). 
Using the results of Tab.~\ref{tab:ablation} (a) (based on  CIFAR-100 and Swin), we chose 
 $m = 64$  for all the VTs and all the datasets.
 Moreover,
 Tab.~\ref{tab:ablation} (b) shows  the influence of the loss weight $\lambda$ (Section~\ref{Method}) for each of  the three baselines,
 which motivates our choice of using $\lambda = 0.1$ for both CvT and T2T and  $\lambda = 0.5$ for Swin.

These values of $m$ and $\lambda$ {\em are kept fixed}
in all the other experiments of this paper, independently of the dataset, the main task (e.g., classification, detection, 
segmentation, etc.), and the training protocol (from scratch or fine-tuning). 
This is done to emphasise the ease of use of our loss.
Finally, in the Supplementary Material, we analyze the influence of the size of the localization MLP ($f$).

\begin{table}[!ht]
	\small 
	\centering
	\caption{CIFAR-100, 100 training epochs: (a) the influence on the  accuracy of the number of pair samples ($m$) in ${L}_{drloc}$ using Swin, and (b) the influence of the $\lambda$ value using all the 3 VT baselines.}
	\renewcommand{\arraystretch}{1.0}
	\subfloat[]{\input{tables/ablationM}}
	\subfloat[]{\input{tables/ablationlambda}}
	\label{tab:ablation}
\end{table}

\begin{table}[!ht]
	\small 
	\centering
	\caption{Top-1 accuracy  on  IN-100 using either 100 or 300 epochs. In the former case, we  show the average and the standard deviation values obtained by repeating each single experiment 5 times with 5 different random seeds. 
	}
	\renewcommand{\arraystretch}{1.0}
	\input{tables/imagenet100}
	\label{tab:results-IN-100}
\end{table}

\subsection{Training from scratch}
\label{scratch}

In this section, we analyze the performance of both the VT baselines and our regularization loss using small-medium  datasets and different number of training epochs, simulating a scenario with limited computational resources and/or limited training data. 
In fact, while fine-tuning a model pre-trained on ImageNet-1K is the most common protocol when dealing with small training datasets, 
this is not possible when, e.g., the
network input is not an RGB image (e.g., in case of  3D point cloud data \cite{PointTransformer})
or when using a task-specific  backbone architecture  \cite{CornerNet,DetNet}.
In these cases, the network needs to be trained from scratch on the target dataset, thus, investigating the robustness of
 the VTs when trained from scratch with relatively small datasets, is useful for those  application domains in which a fine-tuning protocol cannot be adopted.

 We start by analyzing the impact on the accuracy of the number of training epochs on IN-100. Tab.~\ref{tab:results-IN-100} shows that, using  $\mathcal{L}_{drloc}$, {\em all the tested VTs} show an accuracy improvement, and this boost is larger with fewer epochs. As expected, our loss acts as a regularizer, whose effects are more pronounced in a shorter training regime. We believe this result is particularly significant considering the larger computational times which are necessary to train typical VTs with respect to ResNets.

In Tab.~\ref{tab:scratch}, we use all the other datasets and we train from scratch with 100 epochs
(see the Supplementary Material for longer training protocols).
First, we note that the accuracy of the VT baselines varies a lot depending on the dataset (which is expected), but also depending on the specific VT architecture. 
This is largely in contrast with the ImageNet-1K results, where  the difference between the three baselines is much smaller. 
As a reference, when these VTs are trained on ImageNet-1K (for 300 epochs), the differences of their respective top-1 accuracy is much smaller: 
Swin-T,  81.3 \cite{Swin}; T2T-ViT-14, 81.5 \cite{T2T}; CvT-13, 81.6 \cite{CvT}.
Conversely, Tab.~\ref{tab:scratch} shows that, for instance, 
the accuracy difference between CvT and Swin is about 45-46 points in Quickdraw and Sketch, 30 points
on CIFAR-10, and about 20 points on many other datasets.
Analogously, the  difference between CvT and T2T is between 20 and 25 points in Sketch, Painting and Flowers102, and quite significant in the other datasets.
This comparison shows that CvT is usually much more robust in a small training set regime with respect to the other two VTs, a behaviour which is completely hidden when the training/evaluation protocol is based on large datasets only.

In the same table, we also show the accuracy of these three VTs when training is done using $\mathcal{L}_{drloc}$ as a regularizer. 
Similarly to the IN-100 results, also in this case our loss {\em improves the accuracy of all the tested VTs in all the datasets}. Most of the time, this improvement is quite significant (e.g., almost 4 points on SVHN with CvT), and sometimes dramatic (e.g., more than 45 points on Quickdraw with Swin).
These results show that a self-supervised  auxiliary task  can provide a significant "signal" to the VT when the training set is limited, and, specifically, that our loss can be very effective in boosting the accuracy of a VT trained from scratch in this scenario.

In Tab.~\ref{tab:scratch} we also report the results we obtained using a ResNet-50, trained with 100 epochs and the standard ResNet training protocol (e.g., using Mixup \cite{mixup}
and CutMix \cite{CutMix} data-augmentations, etc.). 
These results  show that 
the best performing VT (CvT) is 
usually comparable 
with a same size ResNet,
and demonstrate that VTs can potentially be trained from scratch with darasets smaller than InageNet-1K.
Finally, in the last row of the same table, we train the ResNet-50 baseline jointly with our pretext task. In more detail, we replace the VT token embedding grid ($G_x$ in  Eq.~\ref{eq.drloc-loss}) with the last convolutional feature map of the ResNet, and we apply our loss (Eq.~\ref{eq.drloc-loss}) on top of this  map. A comparison between the results of the last 2 rows of Tab.~\ref{tab:scratch} shows that our loss is useful also when used with a ResNet  (see the Supplementary Material for longer training protocols). 
When using ResNets, the improvement obtained with our loss is marginal, but it is consistent in 9 out of  10 
 datasets. The smaller improvement with respect to the analogous VT results may probably be explained by the fact that ResNets already embed local inductive biases in their architecture, thus a localization auxiliary task is less helpful (Section~\ref{Introduction}).

\begin{table*}[!ht]
\caption{Top-1 accuracy of  VTs and ResNets, trained from scratch on different datasets (100 epochs).}
\centering
\small
\newcolumntype{C}{>{\centering\arraybackslash}X}
\setlength{\tabcolsep}{1pt}
\setlength{\extrarowheight}{5pt}
\renewcommand{\arraystretch}{0.75}
\begin{tabularx}{\linewidth}{Cl!{\color{gray}\vline} CCCC!{\color{gray}\vline}CCCCCC}
	\toprule
	&
	& \rotatebox{90}{\raisebox{0.5pt} CIFAR-10}
	& \rotatebox{90}{\raisebox{0.5pt} CIFAR-100}
	& \rotatebox{90}{\raisebox{0.5pt} Flowers102}
	& \rotatebox{90}{\raisebox{0.5pt} SVHN}
	& \rotatebox{90}{\raisebox{0.5pt} ClipArt}
	& \rotatebox{90}{\raisebox{0.5pt} Infograph}
	& \rotatebox{90}{\raisebox{0.5pt} Painting}
	& \rotatebox{90}{\raisebox{0.5pt} Quickdraw}
	& \rotatebox{90}{\raisebox{0.5pt} Real}
	& \rotatebox{90}{\raisebox{0.5pt} Sketch} \\
	\midrule
	
	\multirow{3}{*}{CvT} & CvT-13 & 89.02 & 73.50 & 54.29 & 91.47 & 60.34 & 19.39 & 54.79 & 70.10 & 76.33 & 56.98\\
	&\CC{20} &\CC{20}\textbf{90.30} &\CC{20}\textbf{74.51} &\CC{20}\textbf{56.29} &\CC{20}\textbf{95.36} &\CC{20} \textbf{60.64}&\CC{20} \textbf{20.05} &\CC{20}\textbf{55.26} &\CC{20}\textbf{70.36} &\CC{20}\textbf{77.05} &\CC{20}\textbf{57.56} \\
	&\CC{20}\multirow{-2}{*}{CvT-13+$\mathcal{L}_{drloc}$} &\CC{20}\posimprov{1.28} &\CC{20}\posimprov{1.01} &\CC{20}\posimprov{2.00} &\CC{20}\posimprov{3.89} &\CC{20}\posimprov{0.30} &\CC{20}\posimprov{0.67} &\CC{20}\posimprov{0.47} &\CC{20}\posimprov{0.26} &\CC{20}\posimprov{0.68} &\CC{20}\posimprov{0.58}\\
	
	\arrayrulecolor{gray}\specialrule{.5pt}{0.6pt}{-0.5pt}\arrayrulecolor{black}
	
	\multirow{3}{*}{Swin} & Swin-T & 59.47 & 53.28 & 34.51 & 71.60 & 38.05 & 8.20 & 35.92 & 24.08 & 73.47 & 11.97\\
	& \CC{20} & \CC{20}\textbf{83.89} & \CC{20}\textbf{66.23} &\CC{20}\textbf{39.37} &\CC{20}\textbf{94.23} & \CC{20}\textbf{47.47} &\CC{20}\textbf{10.16} &\CC{20}\textbf{41.86} &\CC{20}\textbf{69.41} &\CC{20} \textbf{75.59} &\CC{20} \textbf{38.55}  \\
	&\CC{20}\multirow{-2}{*}{Swin-T+$\mathcal{L}_{drloc}$} &\CC{20}\posimprov{24.42} &\CC{20}\posimprov{12.95} &\CC{20}\posimprov{4.86} &\CC{20}\posimprov{22.63} &\CC{20}\posimprov{9.42} &\CC{20}\posimprov{1.96} &\CC{20}\posimprov{5.94} &\CC{20}\posimprov{45.33} &\CC{20}\posimprov{2.12} &\CC{20}\posimprov{26.58}\\
	
	\arrayrulecolor{gray}\specialrule{.5pt}{0.6pt}{-0.5pt}\arrayrulecolor{black}
	
	\multirow{3}{*}{T2T} & T2T-ViT-14 &  84.19 & 65.16 & 31.73 & 95.36 & 43.55 & 6.89 & 34.24 & 69.83 & 73.93 & 31.51 \\
	& \CC{20} &\CC{20}\textbf{87.56} &\CC{20}\textbf{68.03} &\CC{20}\textbf{34.35} &\CC{20}\textbf{96.49} &\CC{20}\textbf{52.36} &\CC{20}\textbf{9.51} &\CC{20}\textbf{42.78} &\CC{20}\textbf{70.16} &\CC{20}\textbf{74.63} &\CC{20} \textbf{51.95} \\
	& \CC{20}\multirow{-2}{*}{T2T-ViT-14+$\mathcal{L}_{drloc}$} &\CC{20}\posimprov{3.37} &\CC{20}\posimprov{2.87} &\CC{20}\posimprov{2.62} &\CC{20}\posimprov{1.13} &\CC{20}\posimprov{8.81} &\CC{20}\posimprov{2.62} &\CC{20}\posimprov{8.54} &\CC{20}\posimprov{0.33} &\CC{20}\posimprov{0.70} &\CC{20}\posimprov{20.44} \\
	
	\arrayrulecolor{gray}\specialrule{.5pt}{0.6pt}{-0.5pt}\arrayrulecolor{black}
	
	\multirow{3}{*}{ResNet} & ResNet-50 & 91.78 & 72.80 & 46.92 & 96.45 & 63.73 & 19.81 & 53.22 & 71.38 & 75.28 & \textbf{60.08} \\
	& \CC{20} & \CC{20}\textbf{92.03} & \CC{20}\textbf{72.94} & \CC{20}\textbf{47.65} & \CC{20}\textbf{96.53} & \CC{20}\textbf{63.93} & \CC{20}\textbf{20.79} & \CC{20}\textbf{53.52} & 
	\CC{20}\textbf{71.57} & \CC{20}\textbf{75.56} & \CC{20}59.62 \\
	& \CC{20}\multirow{-2}{*}{ResNet-50+$\mathcal{L}_{drloc}$} & \CC{20}\posimprov{0.25} & \CC{20}\posimprov{0.14} & \CC{20}\posimprov{0.73} & \CC{20}\posimprov{0.08} & \CC{20}\posimprov{0.20} & \CC{20}\posimprov{0.98} & 
	\CC{20}\posimprov{0.30} & \CC{20}\posimprov{0.19} & \CC{20}\posimprov{0.28} & \CC{20}\negimprov{0.46} \\
	
	\bottomrule
\end{tabularx}

\label{tab:scratch}
\end{table*}
\begin{table*}[!ht]
\caption{Pre-training on ImageNet-1K and then fine-tuning on the target dataset
 (top-1 accuracy,  100 fine-tuning epochs). 
}
\centering
\small
\newcolumntype{C}{>{\centering\arraybackslash}X}
\setlength{\tabcolsep}{1pt}
\setlength{\extrarowheight}{5pt}
\renewcommand{\arraystretch}{0.75}
\begin{tabularx}{\linewidth}{Cl!{\color{gray}\vline} CCCC!{\color{gray}\vline}CCCCCC}
	\toprule
	&
	& \rotatebox{90}{\raisebox{0.5pt} CIFAR-10}
	& \rotatebox{90}{\raisebox{0.5pt} CIFAR-100}
	& \rotatebox{90}{\raisebox{0.5pt} Flowers102}
	& \rotatebox{90}{\raisebox{0.5pt} SVHN}
	& \rotatebox{90}{\raisebox{0.5pt} ClipArt}
	& \rotatebox{90}{\raisebox{0.5pt} Infograph}
	& \rotatebox{90}{\raisebox{0.5pt} Painting}
	& \rotatebox{90}{\raisebox{0.5pt} Quickdraw}
	& \rotatebox{90}{\raisebox{0.5pt} Real}
	& \rotatebox{90}{\raisebox{0.5pt} Sketch} \\
	\midrule
	
	\multirow{3}{*}{Swin} & Swin-T & 97.95 & 88.22 & 98.03 & 96.10 & 73.51 & 41.07 & 72.99 & 75.81 & 85.48 & 72.37\\
	& \CC{20} &\CC{20} \textbf{98.37} &\CC{20} \textbf{88.40} &\CC{20} \textbf{98.21} &\CC{20} \textbf{97.87} &\CC{20} \textbf{79.51} &\CC{20} \textbf{46.10} &\CC{20}\textbf{73.28} &\CC{20}\textbf{76.01} &\CC{20}\textbf{85.61} &\CC{20}\textbf{72.86} \\
	&\CC{20}\multirow{-2}{*}{Swin-T+$\mathcal{L}_{drloc}$} &\CC{20}\posimprov{0.42} &\CC{20}\posimprov{0.18} &\CC{20}\posimprov{0.18} &\CC{20}\posimprov{1.77} &\CC{20}\posimprov{6.00} &\CC{20}\posimprov{5.03} &\CC{20}\posimprov{0.29} &\CC{20}\posimprov{0.20} &\CC{20}\posimprov{0.13} &\CC{20}\posimprov{0.49}\\
	
	\arrayrulecolor{gray}\specialrule{.5pt}{0.6pt}{-0.5pt}\arrayrulecolor{black}
	
	\multirow{3}{*}{T2T} & T2T-ViT-14 & 98.37 & 87.33 & 97.98 & 97.03 & 74.59 & 38.53 & 72.29 & 74.16 & 84.56 & 72.18 \\
	& \CC{20} &\CC{20}\textbf{98.52} &\CC{20}\textbf{87.65} &\CC{20}\textbf{98.08} &\CC{20}\textbf{98.20} &\CC{20}\textbf{78.22} &\CC{20}\textbf{45.69} &\CC{20} \textbf{72.42} &\CC{20}\textbf{74.27} &\CC{20}\textbf{84.57} &\CC{20}\textbf{72.29} \\
	& \CC{20}\multirow{-2}{*}{T2T-ViT-14+$\mathcal{L}_{drloc}$} &\CC{20}\posimprov{0.15} &\CC{20}\posimprov{0.32} &\CC{20}\posimprov{0.10} &\CC{20}\posimprov{1.17} &\CC{20}\posimprov{3.63} &\CC{20}\posimprov{7.16} &\CC{20}\posimprov{0.13} &\CC{20}\posimprov{0.11} &\CC{20}\posimprov{0.01} &\CC{20}\posimprov{0.11} \\
	
	\arrayrulecolor{gray}\specialrule{.5pt}{0.6pt}{-0.5pt}\arrayrulecolor{black}
	
	\multirow{3}{*}{ResNet} & ResNet-50 & 97.65 & 85.44 & 96.59 & 96.60 & 75.22 & 44.30 & 66.58 & 72.12 & 80.40 & 67.77 \\
	& \CC{20} &\CC{20}\textbf{97.74} & \CC{20}\textbf{85.65} & \CC{20}\textbf{96.72} & \CC{20}\textbf{96.71} & \CC{20}\textbf{75.51} & \CC{20}\textbf{44.39} & \CC{20}\textbf{69.03} & \CC{20}\textbf{72.21} & \CC{20}\textbf{80.54} & \CC{20}\textbf{68.14} \\
	& \CC{20}\multirow{-2}{*}{ResNet-50+$\mathcal{L}_{drloc}$} & \CC{20}\posimprov{0.09} & \CC{20}\posimprov{0.21} & \CC{20}\posimprov{0.13} & \CC{20}\posimprov{0.11} & \CC{20}\posimprov{0.29} & \CC{20}\posimprov{0.09} & 
	\CC{20}\posimprov{2.45} & \CC{20}\posimprov{0.09} & \CC{20}\posimprov{0.14} & \CC{20}\posimprov{0.37}\\
	\bottomrule
\end{tabularx}
\label{tab:finetuning}
\vspace{-1em}
\end{table*}


\subsection{Fine-tuning}
\label{finetuning}

In this section, we analyze a typical fine-tuning scenario, in which a model is pre-trained on a big dataset (e.g., ImageNet), and then fine-tuned on the target domain. Specifically, in {\em all} the experiments, we use VT models pre-trained by the corresponding VT authors
on ImageNet-1K {\em without} our localization loss.
The difference between the baselines and ours concerns {\em only} the fine-tuning stage, which is done in the standard way for the former and using 
our $\mathcal{L}_{drloc}$  regularizer for the latter.
Starting from standard pre-trained models and using our loss only in the fine-tuning stage, emphasises the  easy to use of our proposal in practical scenarios, in which fine-tuning can be done without re-training the model on ImageNet.
As mentioned in Section~\ref{sec:experiments}, in this analysis we do not  include CvT because of the lack of publicly available ImageNet-1K pre-trained models for this architecture. 

The results are presented in Tab.~\ref{tab:finetuning}. 
Differently from the results shown in Section~\ref{scratch}, the accuracy
difference between the T2T and Swin baselines is much less  pronounced, and the latter outperforms the former in most of the datasets.
Moreover, analogously to all the other experiments, also in this case, using $\mathcal{L}_{drloc}$ leads to an accuracy improvement {\em with all the tested VTs and in all the datasets}. For instance, on Infograph, Swin with $\mathcal{L}_{drloc}$ improves    of more than 5 points, and  T2T more than 7 points.

In the last two rows of  Tab.~\ref{tab:finetuning}, we show the ResNet based results.
The comparison between ResNet  and the VT baselines shows that the latter  are very competitive in this fine-tuning scenario, 
even more than with a training-from-scratch protocol (Tab.~\ref{tab:scratch}).
For instance, the two VT baselines (without our loss) are
 outperformed by  ResNet only in 2 out of 10 datasets. 
 This confirms that VTs are likely to be widely adopted in computer vision applications in the near future, independently of the training set size.
Finally, analogously to the experiments in Section~\ref{scratch}, 
Tab.~\ref{tab:finetuning} shows that
our loss is (marginally) helpful also  in  ResNet fine-tuning.

%% file: tables/ablationM.tex
\begin{tabular}{l l}
    \toprule
    \setlength{\tabcolsep}{3pt}
    \textbf{Model} & \textbf{Top-1 Acc.} \\
    \midrule
        A: Swin-T & 53.28 \\
        B: A + $\mathcal{L}_{drloc}$, $m$=$32$ & 63.70\\
        C: A + $\mathcal{L}_{drloc}$, $m$=$64$ & \textbf{66.23} \\
        D: A + $\mathcal{L}_{drloc}$, $m$=$128$ & 65.16\\
        E: A + $\mathcal{L}_{drloc}$, $m$=$256$ & 64.87\\
        \bottomrule
    \end{tabular}

%% file: tables/ablationlambda.tex
\begin{tabular}{lllll}
    \toprule
    \setlength{\tabcolsep}{3pt}
    \textbf{Model} & $\lambda$=$0.0$ & $\lambda$=$0.1$ & $\lambda$=$0.5$ & $\lambda$=$1.0$ \\
    \midrule
    CvT-13 & 73.50 & \textbf{74.51} &	74.07 &	72.84\\
    Swin-T & 53.28 & 58.15 & \textbf{66.23} & 64.28\\
    T2T-ViT-14 & 65.16 & \textbf{68.03} &	67.03 &	66.53\\
        \bottomrule
    \end{tabular}

%% file: tables/imagenet100.tex
    \setlength{\extrarowheight}{5pt}
    \renewcommand{\arraystretch}{0.75}
    \small
	\centering
	\begin{tabular}{llcll}
	\toprule
	& \multirow{2}{*}{\textbf{Model}} & \textbf{\# Params} & \multicolumn{2}{c}{\textbf{ImageNet-100}} \\ 
	\cmidrule(lr){4-5} 
	& &\textbf{(M)} & 100 epochs & 300 epochs\\
    \midrule
    \multirow{2}{*}{CvT} & CvT-13 & 20 & 85.62 $\pm$ 0.05 & 90.16 \\
    &\CC{20}CvT-13+$\mathcal{L}_{drloc}$ & \CC{20}20 &\CC{20}\textbf{86.09} $\pm$ 0.12 \posimprov{0.47} &\CC{20}\textbf{90.28} \posimprov{0.12}  \\
    
    \arrayrulecolor{gray}\specialrule{.5pt}{0.6pt}{-0.5pt}\arrayrulecolor{black}
    
    \multirow{2}{*}{Swin} & Swin-T & 29 & 82.66 $\pm$ 0.10 & 89.68\\
    &\CC{20}Swin-T+$\mathcal{L}_{drloc}$& \CC{20}29  &\CC{20}\textbf{83.95} $\pm$ 0.05 \posimprov{1.29} &\CC{20}\textbf{90.32} \posimprov{0.64}\\
    
    \arrayrulecolor{gray}\specialrule{.5pt}{0.6pt}{-0.5pt}\arrayrulecolor{black}
    
	\multirow{2}{*}{T2T} & T2T-ViT-14 & 22 & 82.67 $\pm$ 0.01 &  87.76\\
	&\CC{20}T2T-ViT-14+$\mathcal{L}_{drloc}$& \CC{20}22 &\CC{20}\textbf{83.74} $\pm$ 0.08 \posimprov{1.07} &\CC{20}\textbf{88.16} \posimprov{0.40} \\
	\bottomrule
	\end{tabular}
	

%% file: sections/6-conclusion.tex
\section{Conclusion}
\label{sec:conclusions}

In this paper, we have empirically analyzed different VTs, showing that their performance largely varies when trained  with small-medium  datasets, and that CvT is usually much more effective in generalizing with less data. 
Moreover, we proposed a self-supervised auxiliary task to regularize  VT training.
Our  localization task,  inspired by  \cite{ELECTRA},    is densely defined for  a random subset of final token embedding pairs, and it encourages the VT to learn  spatial information.

In our extensive empirical analysis, with 11 datasets, different training scenarios and three VTs, our dense localization loss {\em has always improved the corresponding baseline accuracy}, usually by a significant margin, and sometimes dramatically (up to +45 points).
We believe that this shows that our proposal is an easy-to-reproduce, yet very effective tool to boost the performance of VTs, especially in training regimes with a limited amount of data/training time. It also paves the way to investigating other forms of self-supervised/multi-task learning which are specific for VTs, and can help VT training without resorting to the use of huge annotated datasets.

{\bf Limitations.}
A deeper analysis on why fine-grained embedding grids have a negative impact on our auxiliary task (Section~\ref{Discussion}) was left as a future work. 
Moreover, despite in the Supplementary Material we show a few experiments with ViT-B, which confirm the usefulness of $\mathcal{L}_{drloc}$ when used with bigger VT models, 
in our analysis we mainly focused on VTs of approximately the same size as a ResNet-50. In fact, the goal of this paper is investigating the VT behaviour with medium-small  datasets, 
thus, high-capacity models  most likely are not the best choice in a training scenario with scarcity of data.

\section*{Acknowledgements}
This work was partially supported by the EU H2020 AI4Media No. 951911 project and by the EUREGIO project OLIVER.

%% file: supplementary/content.tex
\section{Pseudocode of the dense relative localization task}

In order to emphasise the simplicity and the ease of reproduction of our proposed method, 
in Figure~\ref{fig:pytorch-code}  we show  a PyTorch-like pseudocode of  our auxiliary task with the associated $\mathcal{L}_{drloc}$ loss.

\begin{figure}[!ht]
    \input{supplementary/code}
    \caption{A PyTorch-like pseudocode  of our dense relative localization task and the corresponding $\mathcal{L}_{drloc}$ loss.}
     \label{fig:pytorch-code}
\end{figure}

\section{Loss Variants}
\label{Variants}

In  this section, we present different loss function variants associated with our  relative localization task,  which are empirically evaluated in Sec.~\ref{loss-variants-abaltion}.
The goal is to show that the auxiliary task proposed in the main paper can be implemented in different ways and to analyze the differences between these implementations.

The first variant consists in including negative target offsets:
\begin{equation}
	\label{eq.neg.trans-offset}
	t'_u = \frac{i-p}{k}, \quad t'_v = \frac{j-h}{k}, \quad (t'_u,t'_v)^T \in [ -1,  1 ]^2.
\end{equation}
\noindent
Replacing $(t_u,t_v)^T$ in Eq.~\ref{eq.drloc-loss} in the main paper with $(t'_u,t'_v)^T$ computed as in Eq.~\ref{eq.neg.trans-offset}, and keeping all the rest unchanged, we obtain the first variant, which we call $\mathcal{L}_{drloc}^*$.

In the second variant, we transform the regression task in Eq.~\ref{eq.drloc-loss}  in the main paper in a classification task, and we replace the $L_1$ loss with the cross-entropy loss. In more detail, we use as target offsets:
\begin{equation}
	\label{eq.cls.trans-offset}
	c_u = i-p, \quad c_v = j-h, \quad (c_u,c_v)^T \in \{ -k, ..., k \}^2,
\end{equation}
\noindent
and we associate each of the $2 k +1$ discrete elements in $ C = \{ -k, ..., k \}$ with a "class". Accordingly, the localization MLP $f$ is modified by replacing the 2 output neurons with 2 different sets of neurons, one per spatial dimension ($u$ and $v$). Each set of neurons represents a discrete  offset prediction over the $2 k +1$ "classes" in $C$. Softmax is applied {\em separately} to each set of $2 k +1$ neurons, and the output of $f$ is composed of two posterior distributions over $C$: $(\mathbf{p}_u, \mathbf{p}_v)^T = f(\mathbf{e}_{i,j}, \mathbf{e}_{p,h})^T$, where 
$\mathbf{p}_u, \mathbf{p}_v \in [0,1]^{2 k +1}$. Eq.~\ref{eq.drloc-loss} in the main paper is then replaced by:
\begin{equation}
	\label{eq.drloc-loss-ce}
	\mathcal{L}_{drloc}^{ce} = 
	- \sum_{x \in B} 
	\mathbb{E}_{(\mathbf{e}_{i,j}, \mathbf{e}_{p,h}) \sim G_x}
	[ log(\mathbf{p}_u[c_u]) + log(\mathbf{p}_v[c_v]) ],
\end{equation}
\noindent
where $\mathbf{p}_u[c_u]$  indicates the $c_u$-th element of $\mathbf{p}_u$ (and similarly for $\mathbf{p}_v[c_v]$).

Note that, using the cross-entropy loss in Eq.~\ref{eq.drloc-loss-ce}, corresponds to considering $C$ an unordered set of "categories". This implies that prediction errors in $\mathbf{p}_u$ (and $\mathbf{p}_v$) are independent of the "distance" with respect to the ground-truth $c_u$ (respectively, $c_v$).
In order to alleviate this problem, and inspired by \cite{DBLP:conf/cvpr/DwibediATSZ19}, in the third variant we propose, we
impose a Gaussian prior on $\mathbf{p}_u$ and $\mathbf{p}_v$,
and we minimize the normalized squared distance between the expectation of $\mathbf{p}_u$ and the ground-truth $c_u$
(respectively, $\mathbf{p}_v$ and $c_v$). In more detail, let 
$\mu_u = \sum_{c \in C} \mathbf{p}_u[c] * c$ and 
$\sigma_u^2 = \sum_{c \in C} \mathbf{p}_u[c] * (c - \mu_u)^2$ (and similarly for $\mu_v$ and $\sigma_v^2$). Then, Eq.~\ref{eq.drloc-loss-ce} is replaced by:
\begin{equation}
	\label{eq.drloc-loss-ce-reg}
	\mathcal{L}_{drloc}^{reg} = 
	\sum_{x \in B} 
	\mathbb{E}_{(\mathbf{e}_{i,j}, \mathbf{e}_{p,h}) \sim G_x}
	\left[ \frac{(c_u - \mu_u)^2}{\sigma_u^2} + \alpha log(\sigma_u) + 
	\frac{(c_v - \mu_v)^2}{\sigma_v^2} + \alpha log(\sigma_v) \right],
\end{equation}
\noindent
where the terms $log(\sigma_u)$ and $log(\sigma_v)$
are used for variance regularization 
and $\alpha$ weights the importance of the Gaussian prior 
\cite{DBLP:conf/cvpr/DwibediATSZ19}. 
In preliminary experiments in which we tuned the $\alpha$ parameter using Swin, we found that the  default value of $\alpha = 0.001$, as
suggested in \cite{DBLP:conf/cvpr/DwibediATSZ19}, works well in our scenario, thus we adopted it for all the experiments involving  $\mathcal{L}_{drloc}^{reg}$.

The fourth variant we propose is based on a "very-dense" localization loss, where $\mathcal{L}_{drloc}$ is computed {\em for every transformer block} of VT. Specifically, let $G_x^l$ be the $k_l \times k_l$ grid of token embeddings produced by the $l$-th block of the VT, and let $L$ be the total number of these blocks. Then, Eq.~\ref{eq.drloc-loss} in the main paper is replaced by:
\begin{equation}
	\label{eq.drloc-loss-all-layers}
	\mathcal{L}_{drloc}^{all} = 
	\sum_{x \in B} 
	\sum_{l=1}^L
	\mathbb{E}_{(\mathbf{e}_{i,j}, \mathbf{e}_{p,h}) \sim G_x^l}
	[ | (t_u^l,t_v^l)^T - (d_u^l,d_v^l)^T |_1 ],
\end{equation}
\noindent
where $(t_u^l,t_v^l)^T$ and $(d_u^l,d_v^l)^T$ are, respectively,  the target (see main paper Eq.~\ref{eq.trans-offset}) and the prediction offsets computed at block $l$ using the randomly sampled pair $(\mathbf{e}_{i,j}, \mathbf{e}_{p,h}) \in G_x^l$. For each block, we use a block-specific MLP $f^l$ to compute  $(d_u^l,d_v^l)^T$.
Note that, using  Eq.~\ref{eq.drloc-loss-all-layers}, the initial layers of VT receive more "signal", because each block $l$ accumulates the gradients produced by all the  blocks $l' \geq l$.

Apart from $\mathcal{L}_{drloc}^{all}$,
all the other proposed variants 
are very computationally efficient, because they involve only one forward and one backward pass per image, and $m$ forward passes through $f$.

\subsection{Empirical comparison of the loss variants}
\label{loss-variants-abaltion}

In Tab.~\ref{tab:data-ablation}, we compare the loss variants with each other, where the baseline model is Swin \cite{Swin} (row (A)). For these experiments, we use IN-100, we train all the models for 100 epochs, and, as usual, we show the top-1 classification accuracy on the test set.

When we plug $\mathcal{L}_{drloc}$ on top of Swin (main paper, Sec.~\ref{Method}), the final accuracy 
increases  by 1.26 points (B).
All the other dense localization loss variants underperform $\mathcal{L}_{drloc}$ (C-F). A bit surprisingly, the very-dense 
localization loss $\mathcal{L}_{drloc}^{all}$ is significantly outperformed by the much simpler (and computationally more efficient) $\mathcal{L}_{drloc}$. Moreover, $\mathcal{L}_{drloc}^{all}$ is the only variant which underperforms the baseline.
We presume that this is due to the fact that most of the Swin intermediate blocks have   resolution grids $G_x^l$ finer than the last grid 
$G_x^L$ ($l < L$, $k_l > k_L$, Sec.~\ref{Variants}), and this makes the localization task harder, slowing down the convergence of $f^l$,
and likely providing  noisy gradients to the VT (see the discussion in the main paper, Sec.~\ref{Discussion}).
In  all the other experiments (both in the main paper and in this Supplementary Material), we always use $\mathcal{L}_{drloc}$ as the relative localization loss.

\begin{table}[!ht]
	\small 
	\centering
	\caption{IN-100, 100 epoch training: a comparison between  different loss variants.}
	\renewcommand{\arraystretch}{1.0}
	\setlength{\tabcolsep}{6pt}
	\input{tables/ablation}
	\vspace{-1.5em}
	\label{tab:data-ablation}
\end{table}

\subsection{Relative positional embedding}
\label{RelativePositionalEmbedding}

All the loss variants presented in this section have been plugged on Swin, in which relative positional embedding is used (see the main paper, Sec.~\ref{Preliminaries} and Sec.~\ref{Discussion}). However, the  results reported in Tab.~\ref{tab:data-ablation} show that almost all of these losses can boost the accuracy of the Swin baseline. Below, we intuitively explain why 
the relative positional embedding is not sufficient to allow the network to solve our localization task.

The relative positional embedding (called $B$ in \cite{Swin}) used in Swin, is added to the query/key product before the softmax operation (Eq. 4 in \cite{Swin}). The result of this softmax is then used to weight the importance of each "value", and the new embedding representation of each query (i.e., $e_{i,j}$, in our terminology) is given by this weighted sum of values. Thus, the content of $B$ is not directly represented in $e_{i,j}$, but only used to weight the values forming  $e_{i,j}$ (note that there is also a skip connection). For this reason, $B$ may be useful for the task for which it is designed, i.e., computing the importance (attention) of each key with respect to the current query. However, in order to solve our auxiliary task (i.e., to predict  $t_u$ and $t_v$ in 
Eq.~\ref{eq.trans-offset} in the main paper), the VT should be able to recover and extract from a given embedding pair $(e_{i,j}, e_{p,h})$ the specific offset information originally contained in $B_{(i,j), (p,h)}$ and then blended in the value weights. Probably this is a task (much) harder than exploiting appearance information contained in $(e_{i,j}, e_{p,h})$. This is somehow in line with different previous work showing the marginal importance of positional embedding in VTs. For instance, Naseer et al. \cite{DBLP:journals/corr/abs-2105-10497} show that the (absolute) positional embedding used in ViT \cite{ViT}
is {\em not} necessary for the transformer to solve very challenging occlusion or patch permutation tasks, and they conclude that these tasks are solved by ViT thank to its “dynamic receptive field” (i.e., the context represented in each individual token embedding).

\section{Experiments with a larger training budget}
\label{MoreEpochs}

Although the focus of this work is on increasing the VT training efficiency in a scenario with a limited training budget,
in this section we instead investigate the effect of using our auxiliary task on scenarios with a larger training budget. Specifically, 
we test $\mathcal{L}_{drloc}$  with a larger number of training epochs, using higher-capacity VT models and training the VTs on ImageNet-1K. 

In Tab.~\ref{tab:results-imagenet1k} we train both Swin and T2T 
on ImageNet-1K following the standard protocol (e.g., 300 epochs) and using the publicly available code of each VT baseline. 
When we use $\mathcal{L}_{drloc}$, we get a slight improvement with
both 
the baselines, which shows that our loss is beneficial also with larger datasets and longer training schedules (although the margin is smaller with respect to IN-100, see Tab.~\ref{tab:results-IN-100}).

\begin{table}[!ht]
	\small 
	\centering
	\caption{Top-1 accuracy  on ImageNet-1K.
	(*) Results obtained in our run of the  publicly available code with the default hyperparameters of each corresponding VT baseline.}
	\renewcommand{\arraystretch}{1.0}
    \setlength{\extrarowheight}{5pt}
    \renewcommand{\arraystretch}{0.75}
    \small
	\centering
	\begin{tabular}{llll}
	\toprule
	& \textbf{Model} & \textbf{Top-1 Acc.} \\ 
    \midrule
    %
    %
    \multirow{2}{*}{Swin} & Swin-T & 81.2 (*)\\
    &\CC{20}Swin-T+$\mathcal{L}_{drloc}$& \CC{20}\textbf{81.33} \posimprov{0.13}\\
    
    \arrayrulecolor{gray}\specialrule{.5pt}{0.6pt}{-0.5pt}\arrayrulecolor{black}
    
	\multirow{2}{*}{T2T} & T2T-ViT-14  & 80.7 (*)\\
	&\CC{20}T2T-ViT-14+$\mathcal{L}_{drloc}$ &\CC{20}\textbf{80.85} \posimprov{0.15}  \\
	\bottomrule
	\end{tabular}
	\label{tab:results-imagenet1k}
\end{table}

In Tab.~\ref{tab:300epochs-training-from-scratch}, we use the Infograph dataset and we train all the networks 
for 300 epochs. 
The results confirm that $\mathcal{L}_{drloc}$ can improve the final accuracy  
even when a longer training schedule is adopted.
For instance, comparing the results of T2T in Tab.~\ref{tab:300epochs-training-from-scratch} with the T2T results in
Tab.~\ref{tab:scratch} (100 epochs), the relative margin  has significantly increased ($+8.06$  versus  $+2.62$).

\begin{table}[!ht]
	\caption{Infograph, training from scratch with 300 epochs.
	}
	\renewcommand{\arraystretch}{1.1}
	\small 
	\centering
	\label{tab:300epochs-training-from-scratch}
	\begin{tabularx}{0.65\linewidth}{Xll}
		\toprule
		\textbf{Model} & \textbf{Top-1 Acc.} \\
		\midrule
		CvT-13 &	29.76 \\
		\CC{20}CvT-13 + $\mathcal{L}_{drloc}$ &\CC{20}\textbf{30.31} \posimprov{0.55}\\
		\midrule
		Swin-T &	17.17 \\
		\CC{20}Swin-T + $\mathcal{L}_{drloc}$ &\CC{20}\textbf{20.72} \posimprov{3.55}\\
		\midrule
		T2T-ViT-14 	& 12.62 \\
		\CC{20}T2T-ViT-14 + $\mathcal{L}_{drloc}$ &\CC{20}\textbf{20.68} \posimprov{8.06}\\
		\midrule
		ResNet-50 & 29.34 \\
		\CC{20}ResNet-50 + $\mathcal{L}_{drloc}$  & \CC{20}\textbf{30.00} \posimprov{0.66} \\
		\bottomrule
	\end{tabularx}
\end{table}

Finally, in Tab.~\ref{tab:ViT-B}, we use three datasets and  we train from scratch   ViT-B/16 \cite{ViT}, 
which has 86.4 million parameters (about $4\times$ the number of parameters of the other tested VTs and ResNets). 
Note that "16" in ViT-B/16 stands for $16 \times 16$ resolution patches, used as input without patch overlapping.
For a fair comparison, we used for ViT-B/16 the same image resolution ($224 \times 224$) adopted for all the other VTs
(see Sec.~\ref{ImplementationDetails}), thus we get a final ViT-B/16 embedding grid of $14\times14$,
which is pooled to get our $7\times7$ grid as explained in  the main paper (Sec.~\ref{Preliminaries}). For ViT-B/16, we use $\lambda = 0.01$.
Tab.~\ref{tab:ViT-B} shows that our loss 
is effective also with VT models bigger than the three baselines used in the rest of the paper.

\begin{table}[!ht]
	\caption{Training from scratch  ViT-B/16 with 100 epochs.}
	\renewcommand{\arraystretch}{1.1}
	\small 
	\centering
	\label{tab:ViT-B}
	\begin{tabularx}{0.65\linewidth}{Xlll}
		\toprule
		\textbf{Model} & \textbf{CIFAR-10} & \textbf{CIFAR-100} & \textbf{Infograph} \\
		\midrule
		ViT-B/16 &	71.70 &	59.67 &	11.79 \\
		\CC{20}ViT-B/16 + $\mathcal{L}_{drloc}$ &\CC{20}\textbf{73.91} \posimprov{2.21} &\CC{20}\textbf{61.42} \posimprov{1.75} 	&\CC{20}\textbf{12.22} \posimprov{0.43}\\
		\bottomrule
	\end{tabularx}
\end{table}

\section{Transfer to object detection and image segmentation tasks}
\label{AdditionalResults}

In this section, we provide additional fine-tuning experiments using tasks different from classification 
(i.e., object detection, instance segmentation and semantic segmentation). 
Moreover, we use a different training protocol  from the one used in the main paper (Sec.~\ref{finetuning}). Specifically,
the fine-tuning stage is standard ({\em without our loss}), while in the pre-training stage we either use the standard cross-entropy (only), or we pre-train the VT jointly using  the cross-entropy and $\mathcal{L}_{drloc}$.
We adopt the framework proposed in \cite{Swin}, where a pre-trained Swin VT is used as the backbone for detection and segmentation tasks.
In fact, note that Swin is based on a hierarchy of embedding grids, which can be used by the specific object detection/image segmentation architectures as they were convolutional feature maps \cite{Swin}.

The pre-training dataset is either ImageNet-1K or IN-100, 
and in both cases we pre-train Swin using 300 epochs.
Hence, in case of ImageNet-1K pre-training,  the baseline model is fine-tuned starting from 
the Swin-T model corresponding to 
Tab.~\ref{tab:results-imagenet1k}
(final accuracy : 81.2), while Swin-T + $\mathcal{L}_{drloc}$ refers to 
the model trained with our loss in the same table (final accuracy: 81.33). 
Similarly, in case of IN-100 pre-training, the baseline model is fine-tuned starting from 
the Swin-T model corresponding to  
Tab.~\ref{tab:results-IN-100} (final accuracy : 89.68), while Swin-T + $\mathcal{L}_{drloc}$ refers to 
the model trained with our loss in the same table (final accuracy: 90.32). 

The goal of these experiments is to show that the image representation obtained using $\mathcal{L}_{drloc}$ for pre-training, can be usefully transferred to other tasks without modifying the task-specific architecture or the fine-tuning protocol.

\subsection{Object detection and instance segmentation}
\label{sec.detection-segmentation}

\noindent\textbf{Setup.} We strictly follow the  experimental settings used in  Swin~\cite{Swin}.
Specifically, we use  COCO 2017~\cite{lin2014microsoft}, which contains 118K training, 5K validation and 20K test-dev images. We use two popular object detection architectures: Cascade Mask R-CNN~\cite{cai2018cascade} and Mask R-CNN~\cite{he2017mask}, in which the backbone is replaced with the 
pre-trained  Swin model. Moreover, we use the standard  mmcv~\cite{contributors2020mmcv} framework to train and evaluate the models.
 We adopt multi-scale training~\cite{DETR,sun2020sparsercnn} (i.e., we resize the input image such that the shortest side is between 480 and 800 pixels, while the longest side is at most 1333 pixels), the AdamW~\cite{loshchilov2017decoupled} optimizer (initial learning rate  0.0001, weight decay  0.05, and batch size  16), and 
 a 1x schedule (12 epochs with the learning rate decayed by 0.1 at epochs 8 and 11).

\noindent\textbf{Results.} Tab.~\ref{tab:coco-in-1k} shows that Swin-T, pre-trained on ImageNet-1K with  our $\mathcal{L}_{drloc}$ loss, achieves both a higher detection and
a higher instance segmentation accuracy with respect to  the baselines. Specifically, with both  Mask RCNN and Cascade Mask RCNN,
our pre-trained model  outperforms the baselines
with respect to nearly all detection/segmentation metrics.
When pre-training with a smaller dataset (IN-100), the relative improvement is even higher (Tab.~\ref{tab:coco}).

\begin{table}[!ht]
    \caption{{\em ImageNet-1K} pre-training. Results on the COCO object detection and instance segmentation tasks.
    AP$^\text{box}_\text{x}$ and AP$^\text{mask}_\text{x}$ are the standard 
    object detection and segmentation Average Precision metrics, respectively \cite{lin2014microsoft}.
    }
    \renewcommand{\arraystretch}{1.1}
	\small 
	\centering
	\label{tab:coco-in-1k}
	\begin{tabularx}{\linewidth}{Xllllllll}
	\toprule
	\textbf{{Architecture}} & \textbf{Pre-trained backbone} & AP$^\text{box}$ & AP$^\text{box}_\text{50}$ & AP$^\text{box}_\text{75}$ & AP$^\text{mask}$ & AP$^\text{mask}_\text{50}$ & AP$^\text{mask}_\text{75}$ \\
	\midrule
	\multirow{3}{*}{Mask RCNN} & Swin-T & 43.4 & 66.2 & 47.4 & 39.6 & 63.0 & \CC{20}\textbf{42.6} \\
	&\CC{20} &\CC{20}\textbf{43.8} &\CC{20}\textbf{66.5} &\CC{20}\textbf{48.0} &\CC{20}\textbf{39.7} &\CC{20}\textbf{63.1} & 42.5\\
	& \CC{20}\multirow{-2}{*}{Swin-T + $\mathcal{L}_{drloc}$} &\CC{20}\posimprov{0.4} &\CC{20}\posimprov{0.3} &\CC{20}\posimprov{0.6} &\CC{20}\posimprov{0.1} &\CC{20}\posimprov{0.1} &\CC{20}\negimprov{0.1}\\
	\midrule
	\multirow{3}{*}{Cascade Mask RCNN} & Swin-T & 48.0 & 67.1 & 51.7 & 41.5 & 64.3 & \CC{20}\textbf{44.8}\\
	& \CC{20} &\CC{20}\textbf{48.2} &\CC{20}\textbf{67.4} &\CC{20}\textbf{52.1} &\CC{20}\textbf{41.7} &\CC{20}\textbf{64.7} &\CC{20}\textbf{44.8}\\
	& \CC{20}\multirow{-2}{*}{Swin-T + $\mathcal{L}_{drloc}$} &\CC{20}\posimprov{0.2} &\CC{20}\posimprov{0.3} &\CC{20}\posimprov{0.4} &\CC{20}\posimprov{0.2} &\CC{20}\posimprov{0.4} &\CC{20}\posimprov{0.0}\\
	\bottomrule
	\end{tabularx}
\end{table}

\begin{table}[!ht]
    \caption{{\em IN-100} pre-training. Results on the COCO object detection and instance segmentation tasks.
    }
    \renewcommand{\arraystretch}{1.1}
	\small 
	\centering
	\label{tab:coco}
	\begin{tabularx}{\linewidth}{Xllllllll}
	\toprule
	\textbf{{Architecture}} & \textbf{Pre-trained backbone} & AP$^\text{box}$ & AP$^\text{box}_\text{50}$ & AP$^\text{box}_\text{75}$ & AP$^\text{mask}$ & AP$^\text{mask}_\text{50}$ & AP$^\text{mask}_\text{75}$ \\
	\midrule
	\multirow{3}{*}{Mask RCNN} & Swin-T & 41.8 & 60.3 & 45.1 & 36.7 & 57.4 & 39.4 \\
	&\CC{20} &\CC{20}\textbf{42.7} &\CC{20}\textbf{61.3} &\CC{20}\textbf{45.9} &\CC{20}\textbf{37.2} &\CC{20}\textbf{58.4} &\CC{20}\textbf{40.0}\\
	& \CC{20}\multirow{-2}{*}{Swin-T + $\mathcal{L}_{drloc}$} &\CC{20}\posimprov{0.9} &\CC{20}\posimprov{1.0} &\CC{20}\posimprov{0.8} &\CC{20}\posimprov{1.0} &\CC{20}\posimprov{1.0} &\CC{20}\posimprov{0.6}\\
	\midrule
	\multirow{3}{*}{Cascade Mask RCNN} & Swin-T & 36.0 & 58.2 & 38.6 & 33.8 & 55.2 & 35.9 \\
	& \CC{20} &\CC{20}\textbf{37.2} &\CC{20}\textbf{59.4} &\CC{20}\textbf{40.3} &\CC{20}\textbf{34.5} &\CC{20}\textbf{56.2} &\CC{20}\textbf{36.6}\\
	& \CC{20}\multirow{-2}{*}{Swin-T + $\mathcal{L}_{drloc}$} &\CC{20}\posimprov{1.2} &\CC{20}\posimprov{1.2} &\CC{20}\posimprov{1.7} &\CC{20}\posimprov{0.7} &\CC{20}\posimprov{1.0} &\CC{20}\posimprov{0.7}\\
	\bottomrule
	\end{tabularx}
\end{table}

\subsection{Semantic segmentation}

\noindent\textbf{Setup.} 
We again follow the  experimental settings adopted in Swin~\cite{Swin}.
Specifically, for the
semantic segmentation experiments, we use the ADE20K dataset \cite{zhou2019semantic}, which is composed of 150 semantic categories, and contains 20K training, 2K validation and 3K testing images. 
Following \cite{Swin}, we use the popular UperNet~\cite{xiao2018unified} architecture with a  Swin  backbone pre-trained 
either on ImageNet-1K or 
on IN-100 (see above).
We use the implementation released by mmcv~\cite{contributors2020mmcv} to train and evaluate all the models.

 When  fine-tuning,  we used the AdamW~\cite{loshchilov2017decoupled} optimizer with an initial learning rate of $6\times10^{-5}$, a weight decay of 0.01, a scheduler with linear learning-rate decay, and a linear warmup of 1,500 iterations. We fine-tuned all the models on 8 Nvidia V100 32GB GPUs with 2 images per GPU for 160K iterations. We adopt the default data augumentation techniques used for segmentation, namely  random horizontal flipping, random re-scaling with a  [0.5, 2.0]  ratio range and random photometric distortion. We use stochastic depth with ratio $0.2$ for all the models, which are trained with an input of 512$\times$512 pixels.
At inference time, we use a multi-scale testing, with image resolutions which are $\{0.5, 0.75, 1.0, 1.25, 1.5, 1.75 \} \times$ of the  training resolution. 

\noindent\textbf{Results.} The results reported in Tab.~\ref{tab:ade20k-in-1k} show that the models pre-trained on ImageNet-1K with the proposed loss 
{\em always} outperform
the baselines with respect to all the segmentation metrics.
Similarly to Sec.~\ref{sec.detection-segmentation}, when a smaller dataset is used for pre-training (IN-100), the observed relative boost is even higher
(Tab.~\ref{tab:ade20k}).

\begin{table}[!ht]
    \caption{{\em ImageNet-1K} pre-training. Semantic segmentation on the ADE20K dataset (testing on the validation set). 
    mIoU and mAcc refer to the mean Intersection over Union and the mean class Accuracy, respectively. The base architecture is UperNet \cite{xiao2018unified}. 
	}
    \renewcommand{\arraystretch}{1.1}
	\small 
	\centering
	\label{tab:ade20k-in-1k}
	\begin{tabularx}{0.6\linewidth}{Xll}
	\toprule
	\textbf{Pre-trained backbone} & \textbf{mIoU} & \textbf{mAcc} \\
	\midrule
	Swin-T & 43.87 & 55.22 \\
	\CC{20} &\CC{20}\textbf{44.33} &\CC{20}\textbf{55.74} \\
	\CC{20}\multirow{-2}{*}{Swin-T + $\mathcal{L}_{drloc}$} &\CC{20}\posimprov{0.46} &\CC{20}\posimprov{0.52}\\
	\bottomrule
	\end{tabularx}
\end{table}

\begin{table}[!ht]
    \caption{{\em IN-100} pre-training. Semantic segmentation on the ADE20K dataset (testing on the validation set) with a UperNet architecture \cite{xiao2018unified}. 
	}
    \renewcommand{\arraystretch}{1.1}
	\small 
	\centering
	\label{tab:ade20k}
	\begin{tabularx}{0.6\linewidth}{Xll}
	\toprule
	\textbf{Pre-trained backbone} & \textbf{mIoU} & \textbf{mAcc} \\
	\midrule
	Swin-T & 36.93 & 47.76 \\
	\CC{20} &\CC{20}\textbf{37.83} &\CC{20}\textbf{48.69} \\
	\CC{20}\multirow{-2}{*}{Swin-T + $\mathcal{L}_{drloc}$} &\CC{20}\posimprov{0.90} &\CC{20}\posimprov{0.93}\\
	\bottomrule
	\end{tabularx}
\end{table}

\section{Training efficiency}

\begin{figure}[ht]
    \centering
    \includegraphics[width=\linewidth]{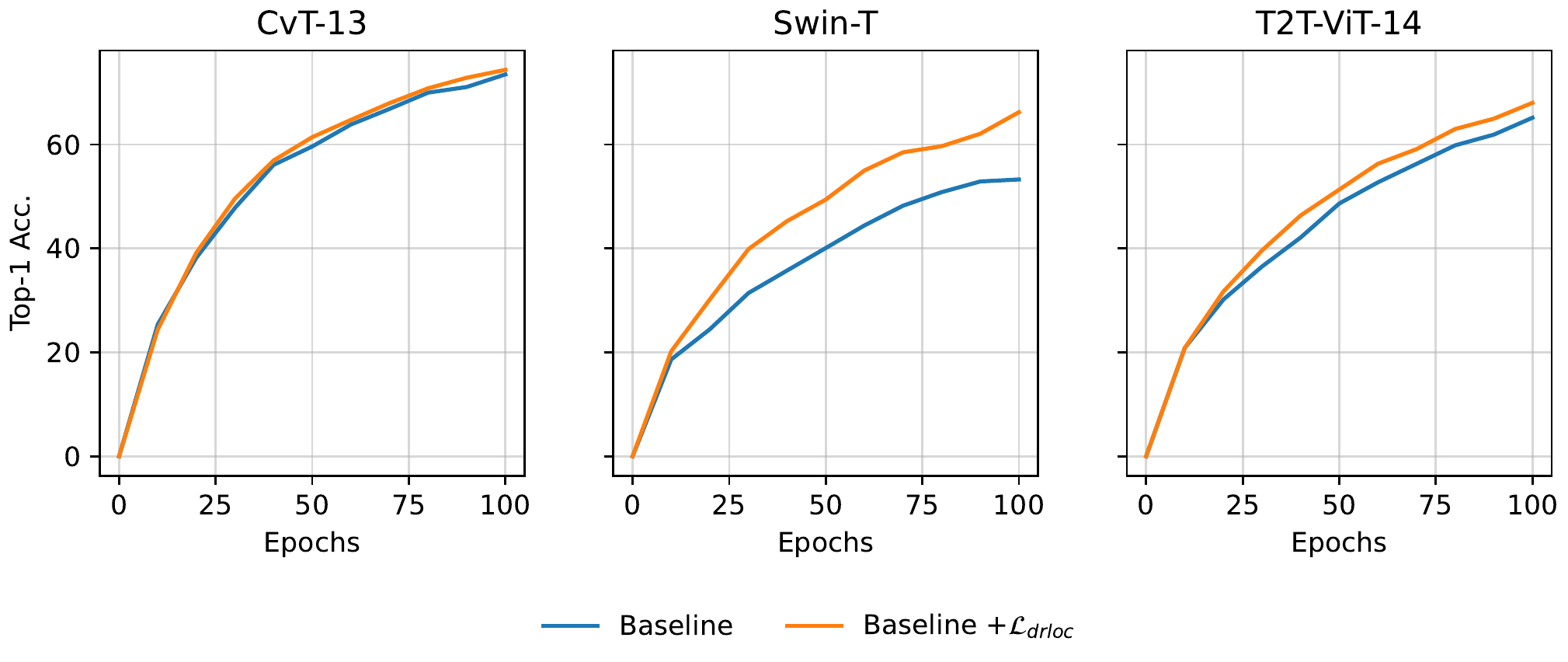}
    \caption{CIFAR-100, training from scratch, top-1 accuracy measured every 10 epochs.
}
    \label{fig:training-curves}
\end{figure}

In Fig.~\ref{fig:training-curves} we show the training curves 
corresponding to the top-1 accuracy of CvT, Swin and T2T, trained from scratch on CIFAR-100, with or without our loss. These graphs show that our auxiliary task is beneficial over 
the whole training stage, and it can speed-up the overall training. For instance, in case of Swin, after 60 training epochs, or method is already significantly better than the  baseline full-trained with 100
epochs ($55.01$ versus $53.28$).

 Finally, we compute the overhead of  $L_{drloc}$ at training time. The results reported in
 Tab.~\ref{tab:efficiency}
 refer to seconds per batch (with a batch size equal to 1024), 
 and show that, overall, the overhead due to our  auxiliary task is negligible with respect to the whole training time.

\begin{table}[!ht]
	\caption{Training time comparison on CIFAR-100. The values are averaged over all training batches  and  jointly reported the corresponding standard deviations.}
	\renewcommand{\arraystretch}{1.1}
	\small 
	\centering
	\label{tab:efficiency}
	\begin{tabularx}{0.6\linewidth}{Xll}
		\toprule
		\textbf{Model} & \textbf{Seconds per batch} \\
		\midrule
		CvT-13 &	0.6037  $\pm$ 0.0040 \\
		\CC{20}CvT-13 + $\mathcal{L}_{drloc}$ &\CC{20}0.6184 $\pm$ 0.0070 \negimprovv{2.43\%}\\
		\midrule
		Swin-T & 0.6684 $\pm$ 0.0031 \\
		\CC{20}Swin-T + $\mathcal{L}_{drloc}$ &\CC{20}0.6842 $\pm$ 0.0033 \negimprovv{2.36\%}\\
		\midrule
		T2T-ViT-14 	& 0.5941 $\pm$ 0.0053 \\
		\CC{20}T2T-ViT-14 + $\mathcal{L}_{drloc}$ &\CC{20}0.6046 $\pm$ 0.0058 \negimprovv{1.77\%}\\
		\bottomrule
	\end{tabularx}
\end{table}

\section{Implementation details and an additional ablation study on the localization MLP}
\label{ImplementationDetails}

Our localization MLP ($f$) is a simple feed-forward network composed of three fully connected 
layers.  The first layer projects the concatenation of the two input token embeddings $\mathbf{e}_{i,j}$ and $\mathbf{e}_{p,h}$ into a 512-dimensional vector and then it applies a \texttt{Relu} activation. Next, we use  a linear layer of dimension 512 followed by a \texttt{Relu} activation. Finally, we use  a linear layer dedicated to the prediction, which depends on the specific loss variant, see  Sec.~\ref{Variants}. For instance, in $\mathcal{L}_{drloc}$, the last layer is composed of two neurons  which predict $d_u$ and $d_v$. 
The details of the MLP head are shown in Tab.~\ref{tab:mlp_head},
while in Tab.~\ref{tab:sizeMLP} we show  the influence of  
 the number of neurons 
in the hidden layers of $f$.

\begin{table}[!ht]
\caption{The details of the localization MLP head. $d$ is the dimension of a token embedding. The number of outputs $o$  and the final nonlinearity (if used) depend on the specific loss. In $\mathcal{L}_{drloc}$, $\mathcal{L}^*_{drloc}$ and $\mathcal{L}_{drloc}^{all}$, 
we use 
$o=2$ without any nonlinearity. Converesely, in both $\mathcal{L}_{drloc}^{ce}$ and $\mathcal{L}_{drloc}^{reg}$, the last layer is split in two branches of $2k+1$ neurons each, and, on each branch, we separately apply a  \texttt{SoftMax} layer.
}
\centering
\label{tab:mlp_head}
\begin{tabular}{lccc}
\toprule
\textbf{Layer} & \textbf{Activation} & \textbf{Output dimension}\\
\midrule 
Input & - & $d$ * 2\\
\midrule 
Linear & ReLU & 512\\
Linear & ReLU & 512\\
Linear & - / SoftMax & $o$\\
\bottomrule
\end{tabular}
\end{table}

\begin{table}[!ht]
	\caption{CIFAR-100, 100 epochs, training from scratch: the influence of the number of neurons used in each of the two hidden layers of the localization MLP.}
	\renewcommand{\arraystretch}{1.1}
	\small 
	\centering
	\label{tab:sizeMLP}
	\begin{tabularx}{0.6\linewidth}{Xlll}
		\toprule
		\textbf{Model} & \multicolumn{3}{c}{\textbf{Number of neurons}} \\
		\cmidrule(lr){2-4}
		& 256 & 512 & 1024 \\
		\midrule
		CvT-13 + $\mathcal{L}_{drloc}$ & 74.19 	& \textbf{74.51} & 	73.80\\
		\midrule
		Swin-T + $\mathcal{L}_{drloc}$ & 65.06 	& \textbf{66.23} & 	64.33\\
		\midrule
		T2T-ViT-14 + $\mathcal{L}_{drloc}$ & 66.49 	& \textbf{68.03} & 	67.83\\
		\bottomrule
	\end{tabularx}
\end{table}

In our experiments, we used  the  officially released framework of Swin~\cite{Swin}\footnote{\url{https://github.com/microsoft/Swin-Transformer}}, which also provides all the necessary code to train and test VT networks (including the object detection and segmentation tasks of Sec.~\ref{AdditionalResults}). For a fair comparison, we use the official  code of T2T-ViT~\cite{T2T}\footnote{\url{https://github.com/yitu-opensource/T2T-ViT}} and a publicly released code of CvT~\cite{CvT}\footnote{\url{https://github.com/lucidrains/vit-pytorch}} and we insert them in the training framework released by the authors of Swin. 
 At submission time of this paper, the official  code of CvT~\cite{CvT} was not publicly available. 
Finally, the ViT-B/16 model used in Sec.~\ref{MoreEpochs} is based on a
public code\footnotemark[\value{footnote}].

When we train the networks from scratch (100 epochs),  we use the AdamW~\cite{loshchilov2017decoupled} optimizer  with a cosine decay learning-rate scheduler and 20 epochs of linear warm-up. We use a batch size of 1024, an initial learning rate of 0.001, and a weight decay of 0.05. When we fine-tune the networks (100 epochs), we use the AdamW~\cite{loshchilov2017decoupled} optimizer  with a cosine decay learning-rate scheduler and 10 epochs of linear warm-up. We use a batch size of 1024, an initial learning rate of 0.0005, and a weight decay of 0.05.
In all the experiments,  the  images of all the datasets are resized to the same fixed resolution ($224 \times 224$).

%% file: supplementary/code.tex
\begin{minted}[
fontsize=\fontsize{8.5}{9},
frame=single,
framesep=2.5pt,
baselinestretch=1.05,
]{python}
# n       : batch size
# m       : number of pairs 
# k X k   : resolution of the embedding grid
# D       : dimension of each token embedding
# x       : a tensor of n embedding grids, shape=[n, D, k, k]

def position_sampling(k, m, n):
    pos_1 = torch.randint(k, size=(n, m, 2))
    pos_2 = torch.randint(k, size=(n, m, 2))
    return pos_1, pos_2

def collect_samples(x, pos, n):
    _, c, h, w = x.size()
    x = x.view(n, c, -1).permute(1, 0, 2).reshape(c, -1)
    pos = ((torch.arange(n).long().to(pos.device) * h * w).view(n, 1) 
        + pos[:, :, 0] * h + pos[:, :, 1]).view(-1) 
    return (x[:, pos]).view(c, n, -1).permute(1, 0, 2)
    
def dense_relative_localization_loss(x):
    n, D, k, k = x.size()
    pos_1, pos_2 = position_sampling(k, m, n)

    deltaxy = abs((pos_1 - pos_2).float()) # [n, m, 2]
    deltaxy /= k

    pts_1 = collect_samples(x, pos_1, n).transpose(1, 2) # [n, m, D]
    pts_2 = collect_samples(x, pos_2, n).transpose(1, 2) # [n, m, D]
    predxy = MLP(torch.cat([pts_1, pts_2], dim=2))
    return L1Loss(predxy, deltaxy)
        
\end{minted}

%% file: tables/ablation.tex
\begin{tabular}{ll ll}
    \toprule
    & \textbf{Model} & \textbf{Top-1 Acc.} \\
    \midrule
        A: & Swin-T & 82.76 \\
        B: & A + $\mathcal{L}_{drloc}$ & 84.02 \posimprov{1.26} \\
        C: & A + $\mathcal{L}^*_{drloc}$ & 83.14 \posimprov{0.38}\\
        D: & A + $\mathcal{L}_{drloc}^{ce}$ & 83.86 \posimprov{1.10}\\
        E: & A + $\mathcal{L}_{drloc}^{reg}$ & 83.24 \posimprov{0.48}\\
        F: & A + $\mathcal{L}_{drloc}^{all}$  & 81.88 \negimprov{0.88} \\
        \bottomrule
    \end{tabular}

%% file: main.bbl
\begin{thebibliography}{10}

\bibitem{DBLP:conf/iclr/AsanoRV20a}
Yuki~Markus Asano, Christian Rupprecht, and Andrea Vedaldi.
\newblock Self-labelling via simultaneous clustering and representation
  learning.
\newblock In {\em ICLR}, 2020.

\bibitem{VICReg}
Adrien Bardes, Jean Ponce, and Yann LeCun.
\newblock {VICReg}: Variance-invariance-covariance regularization for
  self-supervised learning.
\newblock {\em arXiv:2105.04906}, 2021.

\bibitem{NIPS2016_65fc52ed}
Miguel~A Bautista, Artsiom Sanakoyeu, Ekaterina Tikhoncheva, and Bjorn Ommer.
\newblock {CliqueCNN:} deep unsupervised exemplar learning.
\newblock In {\em NeurIPS}, 2016.

\bibitem{cai2018cascade}
Zhaowei Cai and Nuno Vasconcelos.
\newblock Cascade {R-CNN}: Delving into high quality object detection.
\newblock In {\em CVPR}, 2018.

\bibitem{DETR}
Nicolas Carion, Francisco Massa, Gabriel Synnaeve, Nicolas Usunier, Alexander
  Kirillov, and Sergey Zagoruyko.
\newblock End-to-end object detection with transformers.
\newblock In {\em ECCV}, 2020.

\bibitem{DeepClustering}
Mathilde Caron, Piotr Bojanowski, Armand Joulin, and Matthijs Douze.
\newblock Deep clustering for unsupervised learning of visual features.
\newblock In {\em ECCV}, 2018.

\bibitem{SwAV}
Mathilde Caron, Ishan Misra, Julien Mairal, Priya Goyal, Piotr Bojanowski, and
  Armand Joulin.
\newblock Unsupervised learning of visual features by contrasting cluster
  assignments.
\newblock In {\em NeurIPS}, 2020.

\bibitem{DINO}
Mathilde Caron, Hugo Touvron, Ishan Misra, Herv{\'{e}} J{\'{e}}gou, Julien
  Mairal, Piotr Bojanowski, and Armand Joulin.
\newblock Emerging properties in self-supervised vision transformers.
\newblock {\em arXiv:2104.14294}, 2021.

\bibitem{iGPT}
Mark Chen, Alec Radford, Rewon Child, Jeffrey Wu, Heewoo Jun, David Luan, and
  Ilya Sutskever.
\newblock Generative pretraining from pixels.
\newblock In {\em ICML}, 2020.

\bibitem{simclr}
Ting Chen, Simon Kornblith, Mohammad Norouzi, and Geoffrey~E. Hinton.
\newblock A simple framework for contrastive learning of visual
  representations.
\newblock In {\em ICML}, 2020.

\bibitem{simsiam}
Xinlei Chen and Kaiming He.
\newblock Exploring simple siamese representation learning.
\newblock In {\em CVPR}, 2021.

\bibitem{DBLP:journals/corr/abs-2104-02057}
Xinlei Chen, Saining Xie, and Kaiming He.
\newblock An empirical study of training self-supervised vision transformers.
\newblock {\em ICCV}, 2021.

\bibitem{ELECTRA}
Kevin Clark, Minh-Thang Luong, Quoc~V. Le, and Christopher~D. Manning.
\newblock {ELECTRA: P}re-training text encoders as discriminators rather than
  generators.
\newblock In {\em ICLR}, 2020.

\bibitem{contributors2020mmcv}
MMCV Contributors.
\newblock Openmmlab foundational library for computer vision research, 2020.

\bibitem{DBLP:journals/corr/abs-2009-09796}
Michael Crawshaw.
\newblock Multi-task learning with deep neural networks: {A} survey.
\newblock {\em arXiv:2009.09796}, 2020.

\bibitem{UP-DETR}
Zhigang Dai, Bolun Cai, Yugeng Lin, and Junying Chen.
\newblock {UP-DETR:} unsupervised pre-training for object detection with
  transformers.
\newblock In {\em CVPR}, 2021.

\bibitem{devlin-etal-2019-bert}
Jacob Devlin, Ming-Wei Chang, Kenton Lee, and Kristina Toutanova.
\newblock {BERT}: Pre-training of deep bidirectional transformers for language
  understanding.
\newblock In {\em NAACL,}, 2019.

\bibitem{doersch2015unsupervised}
Carl Doersch, Abhinav Gupta, and Alexei~A Efros.
\newblock Unsupervised visual representation learning by context prediction.
\newblock In {\em ICCV}, 2015.

\bibitem{ViT}
Alexey Dosovitskiy, Lucas Beyer, Alexander Kolesnikov, Dirk Weissenborn,
  Xiaohua Zhai, Thomas Unterthiner, Mostafa Dehghani, Matthias Minderer, Georg
  Heigold, Sylvain Gelly, Jakob Uszkoreit, and Neil Houlsby.
\newblock An image is worth 16x16 words: Transformers for image recognition at
  scale.
\newblock In {\em ICLR}, 2021.

\bibitem{DBLP:conf/cvpr/DwibediATSZ19}
Debidatta Dwibedi, Yusuf Aytar, Jonathan Tompson, Pierre Sermanet, and Andrew
  Zisserman.
\newblock Temporal cycle-consistency learning.
\newblock In {\em CVPR}, 2019.

\bibitem{little-friends}
Debidatta Dwibedi, Yusuf Aytar, Jonathan Tompson, Pierre Sermanet, and Andrew
  Zisserman.
\newblock With a little help from my friends: Nearest-neighbor contrastive
  learning of visual representations.
\newblock In {\em ICCV}, 2021.

\bibitem{w-mse}
Aleksandr Ermolov, Aliaksandr Siarohin, Enver Sangineto, and Nicu Sebe.
\newblock Whitening for self-supervised representation learning.
\newblock In {\em ICML}, 2021.

\bibitem{DBLP:conf/eccv/GansbekeVGPG20}
Wouter~Van Gansbeke, Simon Vandenhende, Stamatios Georgoulis, Marc Proesmans,
  and Luc~Van Gool.
\newblock {SCAN:} learning to classify images without labels.
\newblock In {\em ECCV}, 2020.

\bibitem{RotNet}
Spyros Gidaris, Praveer Singh, and Nikos Komodakis.
\newblock Unsupervised representation learning by predicting image rotations.
\newblock In {\em ICLR}, 2018.

\bibitem{byol}
Jean-Bastien Grill, Florian Strub, Florent Altché, Corentin Tallec, Pierre~H.
  Richemond, Elena Buchatskaya, Carl Doersch, Bernardo~Avila Pires,
  Zhaohan~Daniel Guo, Mohammad~Gheshlaghi Azar, Bilal Piot, Koray Kavukcuoglu,
  Rémi Munos, and Michal Valko.
\newblock Bootstrap your own latent: A new approach to self-supervised
  learning.
\newblock In {\em NeurIPS}, 2020.

\bibitem{MoCo}
Kaiming He, Haoqi Fan, Yuxin Wu, Saining Xie, and Ross Girshick.
\newblock Momentum contrast for unsupervised visual representation learning.
\newblock In {\em CVPR}, 2020.

\bibitem{he2017mask}
Kaiming He, Georgia Gkioxari, Piotr Doll{\'a}r, and Ross Girshick.
\newblock Mask {R-CNN}.
\newblock In {\em ICCV}, 2017.

\bibitem{he2016deep}
Kaiming He, Xiangyu Zhang, Shaoqing Ren, and Jian Sun.
\newblock Deep residual learning for image recognition.
\newblock In {\em CVPR}, 2016.

\bibitem{DIM}
R.~Devon Hjelm, Alex Fedorov, Samuel Lavoie{-}Marchildon, Karan Grewal, Philip
  Bachman, Adam Trischler, and Yoshua Bengio.
\newblock Learning deep representations by mutual information estimation and
  maximization.
\newblock In {\em ICLR}, 2019.

\bibitem{DBLP:conf/iccv/HuZXL19}
Han Hu, Zheng Zhang, Zhenda Xie, and Stephen Lin.
\newblock Local relation networks for image recognition.
\newblock In {\em ICCV}, 2019.

\bibitem{hua2021feature}
Tianyu Hua, Wenxiao Wang, Zihui Xue, Yue Wang, Sucheng Ren, and Hang Zhao.
\newblock On feature decorrelation in self-supervised learning.
\newblock {\em arXiv:2105.00470}, 2021.

\bibitem{GAT}
Drew~A. Hudson and C.~Lawrence Zitnick.
\newblock {Generative Adversarial Transformers}.
\newblock In {\em ICML}, 2021.

\bibitem{IIC}
Xu~Ji, Jo{\~{a}}o~F. Henriques, and Andrea Vedaldi.
\newblock Invariant information clustering for unsupervised image
  classification and segmentation.
\newblock In {\em ICCV}, 2019.

\bibitem{TransGAN}
Yifan Jiang, Shiyu Chang, and Zhangyang Wang.
\newblock {TransGAN}: Two transformers can make one strong {GAN}.
\newblock {\em arXiv:2102.07074}, 2021.

\bibitem{cifar}
Alex Krizhevsky, Geoffrey Hinton, et~al.
\newblock Learning multiple layers of features from tiny images.
\newblock 2009.

\bibitem{CornerNet}
Hei Law and Jia Deng.
\newblock Cornernet: Detecting objects as paired keypoints.
\newblock {\em Int. J. Comput. Vis.}, 128(3):642--656, 2020.

\bibitem{LocalViT}
Yawei Li, Kai Zhang, Jiezhang Cao, Radu Timofte, and Luc~Van Gool.
\newblock {LocalViT}: Bringing locality to vision transformers.
\newblock {\em arXiv:2104.05707}, 2021.

\bibitem{DetNet}
Zeming Li, Chao Peng, Gang Yu, Xiangyu Zhang, Yangdong Deng, and Jian Sun.
\newblock Detnet: Design backbone for object detection.
\newblock In {\em ECCV}, 2018.

\bibitem{lin2014microsoft}
Tsung-Yi Lin, Michael Maire, Serge Belongie, James Hays, Pietro Perona, Deva
  Ramanan, Piotr Doll{\'a}r, and C~Lawrence Zitnick.
\newblock Microsoft {COCO}: Common objects in context.
\newblock In {\em ECCV}, 2014.

\bibitem{Swin}
Ze~Liu, Yutong Lin, Yue Cao, Han Hu, Yixuan Wei, Zheng Zhang, Stephen Lin, and
  Baining Guo.
\newblock Swin transformer: Hierarchical vision transformer using shifted
  windows.
\newblock {\em arXiv:2103.14030}, 2021.

\bibitem{loshchilov2017decoupled}
Ilya Loshchilov and Frank Hutter.
\newblock Decoupled weight decay regularization.
\newblock {\em arXiv:1711.05101}, 2017.

\bibitem{TrackFormer}
Tim Meinhardt, Alexander Kirillov, Laura Leal{-}Taix{\'{e}}, and Christoph
  Feichtenhofer.
\newblock {TrackFormer}: Multi-object tracking with transformers.
\newblock {\em arXiv:2101.02702}, 2021.

\bibitem{DBLP:journals/corr/abs-1301-3781}
Tomas Mikolov, Kai Chen, Greg Corrado, and Jeffrey Dean.
\newblock Efficient estimation of word representations in vector space.
\newblock {\em arXiv:1301.3781}, 2013.

\bibitem{DBLP:conf/nips/MikolovSCCD13}
Tomas Mikolov, Ilya Sutskever, Kai Chen, Gregory~S. Corrado, and Jeffrey Dean.
\newblock Distributed representations of words and phrases and their
  compositionality.
\newblock In {\em NeurIPS}, 2013.

\bibitem{misra2019selfsupervised}
Ishan Misra and Laurens van~der Maaten.
\newblock Self-supervised learning of pretext-invariant representations.
\newblock In {\em CVPR}, 2020.

\bibitem{DBLP:journals/corr/abs-2105-10497}
Muzammal Naseer, Kanchana Ranasinghe, Salman~H. Khan, Munawar Hayat,
  Fahad~Shahbaz Khan, and Ming{-}Hsuan Yang.
\newblock Intriguing properties of vision transformers.
\newblock {\em arXiv:2105.10497}, 2021.

\bibitem{netzer2011reading}
Yuval Netzer, Tao Wang, Adam Coates, Alessandro Bissacco, Bo~Wu, and Andrew~Y
  Ng.
\newblock Reading digits in natural images with unsupervised feature learning.
\newblock In {\em NeurIPS Workshop on Deep Learning and Unsupervised Feature
  Learning}, 2011.

\bibitem{flower}
Maria-Elena Nilsback and Andrew Zisserman.
\newblock Automated flower classification over a large number of classes.
\newblock In {\em Indian Conference on Computer Vision, Graphics \& Image
  Processing}, 2008.

\bibitem{noroozi2016unsupervised}
Mehdi Noroozi and Paolo Favaro.
\newblock Unsupervised learning of visual representations by solving jigsaw
  puzzles.
\newblock In {\em ECCV}, 2016.

\bibitem{noroozi2017representation}
Mehdi Noroozi, Hamed Pirsiavash, and Paolo Favaro.
\newblock Representation learning by learning to count.
\newblock In {\em ICCV}, 2017.

\bibitem{peng2019moment}
Xingchao Peng, Qinxun Bai, Xide Xia, Zijun Huang, Kate Saenko, and Bo~Wang.
\newblock Moment matching for multi-source domain adaptation.
\newblock In {\em CVPR}, 2019.

\bibitem{Radford2018ImprovingLU}
Alec Radford and Karthik Narasimhan.
\newblock Improving language understanding by generative pre-training.
\newblock 2018.

\bibitem{DBLP:journals/jmlr/RaffelSRLNMZLL20}
Colin Raffel, Noam Shazeer, Adam Roberts, Katherine Lee, Sharan Narang, Michael
  Matena, Yanqi Zhou, Wei Li, and Peter~J. Liu.
\newblock Exploring the limits of transfer learning with a unified text-to-text
  transformer.
\newblock {\em Journal of Machine Learning Research}, 21:140:1--140:67, 2020.

\bibitem{DBLP:journals/corr/abs-2108-08810}
Maithra Raghu, Thomas Unterthiner, Simon Kornblith, Chiyuan Zhang, and Alexey
  Dosovitskiy.
\newblock Do vision transformers see like convolutional neural networks?
\newblock {\em arXiv:2108.08810}, 2021.

\bibitem{imagenet}
Olga Russakovsky, Jia Deng, Hao Su, Jonathan Krause, Sanjeev Satheesh, Sean Ma,
  Zhiheng Huang, Andrej Karpathy, Aditya Khosla, Michael Bernstein, et~al.
\newblock Imagenet large scale visual recognition challenge.
\newblock {\em International Journal of Computer Vision}, 115(3):211--252,
  2015.

\bibitem{DBLP:conf/naacl/ShawUV18}
Peter Shaw, Jakob Uszkoreit, and Ashish Vaswani.
\newblock Self-attention with relative position representations.
\newblock In {\em NAACL}, 2018.

\bibitem{strudel2021segmenter}
Robin Strudel, Ricardo Garcia, Ivan Laptev, and Cordelia Schmid.
\newblock Segmenter: Transformer for semantic segmentation.
\newblock In {\em ICCV}, 2021.

\bibitem{VideoBERT}
Chen Sun, Austin Myers, Carl Vondrick, Kevin Murphy, and Cordelia Schmid.
\newblock {VideoBERT:} {A} joint model for video and language representation
  learning.
\newblock In {\em ICCV}, 2019.

\bibitem{sun2020sparsercnn}
Peize Sun, Rufeng Zhang, Yi~Jiang, Tao Kong, Chenfeng Xu, Wei Zhan, Masayoshi
  Tomizuka, Lei Li, Zehuan Yuan, Changhu Wang, et~al.
\newblock {Sparse R-CNN}: End-to-end object detection with learnable proposals.
\newblock {\em arXiv:2011.12450}, 2020.

\bibitem{tian2019contrastive}
Yonglong Tian, Dilip Krishnan, and Phillip Isola.
\newblock Contrastive multiview coding.
\newblock In {\em ECCV}, 2020.

\bibitem{DeiT}
Hugo Touvron, Matthieu Cord, Matthijs Douze, Francisco Massa, Alexandre
  Sablayrolles, and Herv{\'e} J{\'e}gou.
\newblock Training data-efficient image transformers \& distillation through
  attention.
\newblock {\em arXiv:2012.12877}, 2020.

\bibitem{CPC}
A{\"{a}}ron van~den Oord, Yazhe Li, and Oriol Vinyals.
\newblock Representation learning with contrastive predictive coding.
\newblock {\em arXiv:1807.03748}, 2018.

\bibitem{attention-is-all-you-need}
Ashish Vaswani, Noam Shazeer, Niki Parmar, Jakob Uszkoreit, Llion Jones,
  Aidan~N. Gomez, Lukasz Kaiser, and Illia Polosukhin.
\newblock Attention is all you need.
\newblock In {\em NeurIPS}, 2017.

\bibitem{wang2020understanding}
Tongzhou Wang and Phillip Isola.
\newblock Understanding contrastive representation learning through alignment
  and uniformity on the hypersphere.
\newblock In {\em ICML}, 2020.

\bibitem{non-local-net}
Xiaolong Wang, Ross~B. Girshick, Abhinav Gupta, and Kaiming He.
\newblock Non-local neural networks.
\newblock In {\em CVPR}, 2018.

\bibitem{CvT}
Haiping Wu, Bin Xiao, Noel Codella, Mengchen Liu, Xiyang Dai, Lu~Yuan, and Lei
  Zhang.
\newblock {CvT}: Introducing convolutions to vision transformers.
\newblock {\em arXiv:2103.15808}, 2021.

\bibitem{xiao2018unified}
Tete Xiao, Yingcheng Liu, Bolei Zhou, Yuning Jiang, and Jian Sun.
\newblock Unified perceptual parsing for scene understanding.
\newblock In {\em ECCV}, 2018.

\bibitem{CoaT}
Weijian Xu, Yifan Xu, Tyler Chang, and Zhuowen Tu.
\newblock Co-scale conv-attentional image transformers.
\newblock {\em arXiv:2104.06399}, 2021.

\bibitem{CeiT}
Kun Yuan, Shaopeng Guo, Ziwei Liu, Aojun Zhou, Fengwei Yu, and Wei Wu.
\newblock Incorporating convolution designs into visual transformers.
\newblock {\em arXiv:2103.11816}, 2021.

\bibitem{T2T}
Li~Yuan, Yunpeng Chen, Tao Wang, Weihao Yu, Yujun Shi, Zihang Jiang, Francis~EH
  Tay, Jiashi Feng, and Shuicheng Yan.
\newblock Tokens-to-token {ViT}: Training vision transformers from scratch on
  {ImageNet}.
\newblock In {\em ICCV}, 2021.

\bibitem{CutMix}
Sangdoo Yun, Dongyoon Han, Sanghyuk Chun, Seong~Joon Oh, Youngjoon Yoo, and
  Junsuk Choe.
\newblock {CutMix: R}egularization strategy to train strong classifiers with
  localizable features.
\newblock In {\em ICCV}, 2019.

\bibitem{barlow}
Jure Zbontar, Li~Jing, Ishan Misra, Yann LeCun, and St{\'{e}}phane Deny.
\newblock Barlow twins: Self-supervised learning via redundancy reduction.
\newblock In {\em ICML}, 2021.

\bibitem{mixup}
Hongyi Zhang, Moustapha Ciss{\'{e}}, Yann~N. Dauphin, and David Lopez{-}Paz.
\newblock mixup: Beyond empirical risk minimization.
\newblock In {\em ICLR}, 2018.

\bibitem{PointTransformer}
Hengshuang Zhao, Li~Jiang, Jiaya Jia, Philip H.~S. Torr, and Vladlen Koltun.
\newblock Point transformer.
\newblock {\em arXiv:2012.09164}, 2020.

\bibitem{zhou2019semantic}
Bolei Zhou, Hang Zhao, Xavier Puig, Tete Xiao, Sanja Fidler, Adela Barriuso,
  and Antonio Torralba.
\newblock Semantic understanding of scenes through the ade20k dataset.
\newblock {\em International Journal of Computer Vision}, 127(3):302--321,
  2019.

\bibitem{Deformable-DETR}
Xizhou Zhu, Weijie Su, Lewei Lu, Bin Li, Xiaogang Wang, and Jifeng Dai.
\newblock Deformable {DETR}: Deformable transformers for end-to-end object
  detection.
\newblock In {\em ICLR}, 2021.

\bibitem{DBLP:journals/corr/abs-1903-12355}
Chengxu Zhuang, Alex~Lin Zhai, and Daniel Yamins.
\newblock Local aggregation for unsupervised learning of visual embeddings.
\newblock In {\em ICCV}, 2019.

\end{thebibliography}
